\documentclass[review]{elsarticle}

\usepackage{lineno,hyperref}
\usepackage{graphicx}
\usepackage{stackengine}
\usepackage{amsfonts}
\usepackage{amsmath}
\usepackage{nccmath}
\usepackage[algo2e,ruled,vlined]{algorithm2e}
\usepackage{multirow}
\usepackage[nice]{nicefrac}
\usepackage{verbatim}

\usepackage{caption}
\usepackage{subfigure}

\usepackage{floatrow}
\usepackage{natbib}

\usepackage{epstopdf}

\usepackage{xcolor}
\usepackage[export]{adjustbox}
\usepackage{booktabs}

\AppendGraphicsExtensions{.tif}

\newcommand*{\LargerCdot}{\raisebox{-0.45ex}{\scalebox{1.4}{$\cdot$}}}

\renewcommand{\vec}[1]{\mathbf{#1}}

\newcommand{\mr}[1]{\mathrm{#1}}
\newcommand{\tx}[1]{\textrm{#1}}
\newcommand{\tr}[1]{#1^\top}

\newcommand{\Real}{\mathbb{R}}

\newcommand{\Reg}{\mathcal{R}}

\newcommand{\ZZ}{\vec Z}
\newcommand{\XX}{\vec X}
\newcommand{\DD}{\vec D}
\newcommand{\xx}{\vec x}
\newcommand{\PPhi}{\vec \Phi}
\newcommand{\AAA}{\vec A}
\newcommand{\KK}{\vec K}
\newcommand{\WW}{\vec W}
\newcommand{\ww}{\vec w}

\newcommand{\pphi}{{\vec \phi}}

\newcommand{\LLL}{\vec L}

\newcommand{\LL}{\mathcal{L}}

\newcommand{\II}{\vec I}
\newcommand{\UU}{\vec U}

\newcommand{\PP}{\vec P}
\newcommand{\QQ}{\vec Q}
\newcommand{\RR}{\vec R}

\newcommand{\GG}{\vec G}

\newcommand{\kk}{\vec k}

\newcommand{\vo}{\vec 1}

\newcommand{\zz}{\vec z}

\newcommand{\wri}{\ww_{i\LargerCdot}}

\newcommand{\zri}{\zz_{i\LargerCdot}}
\newcommand{\zhri}{\hat{\zz}_{i\LargerCdot}}

\DeclareMathOperator*{\argmin}{arg\,min}

\newcommand{\beq}{\begin{equation}}
	\newcommand{\eeq}{\end{equation}}
\newcommand{\beqa}{\begin{eqnarray}}
	\newcommand{\eeqa}{\end{eqnarray}}
\newcommand{\beqas}{\begin{eqnarray*}}
	\newcommand{\eeqas}{\end{eqnarray*}}

\begin{document}
	
	\date{}
	
	\begin{frontmatter}
		
		\title{White matter fiber analysis using kernel dictionary learning and sparsity priors}
		
		%
		%
		%

		\author{Kuldeep Kumar$^{a,*}$\fnref{myfootnote}}
		\author{Kaleem Siddiqi$^{b}$, Christian Desrosiers$^{a}$}
		\fntext[myfootnote]{kkumar@livia.etsmtl.ca}
		
		\address[a]{Laboratory for Imagery, Vision and Artificial Intelligence, \'Ecole de technologie sup\'erieure, 1100 Notre-Dame W., Montreal, QC, Canada, H3C1K3}
		\address[b]{School of Computer Science \& Center for Intelligent Machines, McGill University, \\ 3480 University Street, Montreal, QC, Canada, H3A2A7}
		
		\begin{abstract}
			Diffusion magnetic resonance imaging, a non-invasive tool to infer white matter fiber connections, produces a large number of streamlines containing a wealth of information on structural connectivity. The size of these tractography outputs makes further analyses complex, creating a need for methods to group streamlines into meaningful bundles. In this work, we address this by proposing a set of kernel dictionary learning and sparsity priors based methods. Proposed frameworks include $L_0$ norm, group sparsity, as well as manifold regularization prior. The proposed methods allow streamlines to be assigned to more than one bundle, making it more robust to overlapping bundles and inter-subject variations. We evaluate the performance of our method on a labeled set and data from Human Connectome Project. Results highlight the ability of our method to group streamlines into plausible bundles and illustrate the impact of sparsity priors on the performance of the proposed methods.
		\end{abstract}
		
		
	\end{frontmatter}
	
	
	
	\section{Introduction}
	\label{sec:Intro}
	
	Since its development in the 1980s, diffusion tensor imaging (DTI) has become an essential tool to study white matter connectivity in the human brain. Its ability to infer the orientation of white matter fibers, in-vivo and non-invasively, is key to understanding brain connectivity and associated	neurological diseases \cite{hagmann2006understanding,de2011atlasing}. Since the macroscopic inference of underlying fibers from dMRI data, known as tractography, typically produces a large number of streamlines, it is common to group these streamlines into anatomically meaningful clusters called \emph{bundles} \cite{o2013fiber}. Clustering streamlines is also essential for the creation of white matter atlases, visualization, and statistical analysis of microstructure measures along tracts \cite{guevara2012automatic,Odonnell07automatictractography,siless2018anatomicuts}. 
	Furthermore, clinical applications of tractography analysis are also numerous and include identifying major bundles for neurological planning in patients with tumors \cite{o2017automated}, understanding difference between white matter connectivity in typically developing controls versus children with autism \cite{zhang2017whole}, and uncovering white matter bundles as bio-markers for the diagnosis of Parkinson's disease \cite{cousineau2017test}. 
	
\begin{figure}[h]
		{\centering
			
			\includegraphics{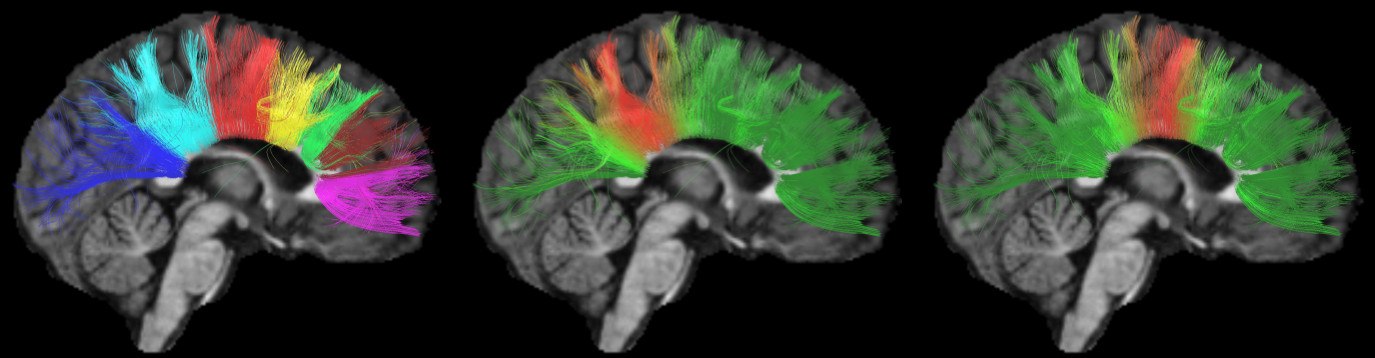}
			
		}
		\caption{Illustrative example. Clustering of the corpus callosum by our method: hard clustering (\textbf{left}), and membership of each streamline to two bundles (\textbf{center} and \textbf{right}). Dark green represents a zero membership and bright red a maximum membership to the bundles.}
		\label{fig:corpus-callosum_KSC}
	\end{figure}

	Clustering streamlines into anatomically meaningful bundles is a challenging task in part due to lack of gold standard. There can be several hundreds of thousands of streamlines to consider, making the clustering problem computationally complex. As illustrated in Fig. \ref{fig:corpus-callosum_KSC}, streamlines within the same bundle can have different lengths and endpoints. Thus, using standard geometric distance measures often leads to poor results. Another challenge comes from the weak separability of certain bundles, which can result in low-quality (e.g., too small or too large) clusters. Also, while many clustering approaches assume a crisp membership of streamlines to bundles, as shown in Fig. \ref{fig:corpus-callosum_KSC}, such a separation of streamlines into hard clusters is often arbitrary. In practice, streamline bundles may overlap and intersect each other, making their extraction and analysis difficult. 
	Moreover, when used to label the streamlines of a new subject, the clusters learned using crisp methods often give unsatisfactory results due to the variability across individual brains. 
	
	In this paper, we propose a set of flexible and efficient streamline clustering approaches based on kernel dictionary learning and sparsity priors. The general idea of these approaches is to learn a compact dictionary of training streamlines capable of describing the whole dataset, and to encode bundles as a sparse non-negative combination of multiple dictionary prototypes. In contrast to spectral embedding methods (e.g., \cite{brun2004clustering,o2005white}) which perform the embedding and clustering in two separate steps, our approaches find clusters in the kernel space without having to explicitly compute an embedding. 
	
	The proposed streamline clustering approaches have several advantages over existing methods for this task. First, they do not require an explicit representation of the streamlines and can extend to any streamline representation or distance/similarity measure. Second, they use a non-linear kernel mapping which facilitates the separation of clusters in a manifold space. Third, unlike hard-clustering methods like the k-means algorithm and its variants (e.g. spectral clustering), they can distribute the membership of streamlines across multiple bundles, making them more robust to overlapping bundles and outliers, as well as to variability across subjects. 
	
	Our specific contributions include: 
	\begin{enumerate}
		\setlength{\itemsep}{2pt}%
		\setlength{\parskip}{2pt}
		\item We propose three different streamline clustering models based on kernel k-means, non-negative factorization and sparse coding, and demonstrate the advantages of these models with respect to the state of the art;
		
		\item We provide a flexible platform to integrate and evaluate streamline distance measures, and compare the performance of three popular measures using two different datasets;
		
		\item Whereas dictionary learning and sparsity have shown promise in various pattern recognition and neuroimaging applications, to our knowledge, the present article is the first account of their use for streamline clustering in a peer-reviewed indexed publication.
		Our results on the streamline clustering problem show the potential of this approach for other imaging applications.
	\end{enumerate}

	The rest of the paper is structured as follows. Section \ref{sec:related-works} provides a brief survey of relevant literature on streamline clustering. In Section \ref{sec:MaterialsAndMethods}, we present our kernel dictionary learning based methods. Section \ref{sec:Results} evaluates the methods on the task of clustering streamlines using real data. Finally, we conclude with a summary of our main contributions, and discuss potential extensions.

	\section{Related works}\label{sec:related-works}
	
	Our presentation of relevant work is divided into two parts, focusing respectively on the various approaches for representation and analysis of streamlines, and the application of sparse coding techniques in neuroimaging.
	
	\subsection{White matter fiber analysis}
	
	
	Over the years, several approaches have been proposed to cluster streamlines and provide a simplified quantitative description of white matter connections, including cross-population inferences \cite{guevara2012automatic,jin2014automatic,o2007automatic,prasad2014automatic}. These studies could be vaguely classified into two categories: representation of streamlines or streamline similarity, and clustering approaches. Features proposed to represent streamlines include the distribution parameters (mean and covariance) of points along the streamline \cite{brun2004clustering} and B-splines \cite{maddah2006statistical}. Approaches using such explicit features typically suffer from two problems: they are sensitive to the length and endpoint positions of the streamlines  and/or are unable to capture their full shape. Instead of using explicit features, streamlines can also be compared using specialized distance measures. Popular distance measures for this task include the Hausdorff distance, the Minimum Direct Flip (MDF) distance and the Mean Closest Points (MCP) distance \cite{corouge2004towards,moberts2005evaluation}. 
	
	Fiber clustering approaches include manifold embedding techniques such as spectral clustering and normalized cuts \cite{brun2004clustering}, agglomerative approaches like hierarchical clustering \cite{Odonnell07automatictractography,corouge2004towards}, k-means \cite{li2010hybrid}, and Dirichlet processes
	\cite{wassermann2010unsupervised,wang2011tractography}. Several studies have also focused on incorporating anatomical features into the clustering \cite{siless2018anatomicuts,o2007automatic}, or on clustering large multi-subject datasets \cite{guevara2012automatic}. A detailed description and comparison of several distances and clustering approaches can be found in \cite{moberts2005evaluation,olivetti2017comparison,siless2013comparison}.
	
	Various studies have also focused on the segmentation of streamlines, toward the goal of drawing cross-population inferences \cite{guevara2012automatic,jin2014automatic,o2007automatic,prasad2014automatic}. These studies either follow an atlas based approach \cite{guevara2012automatic,jin2014automatic,o2007automatic} or align specific tracts directly across subjects \cite{garyfallidis2015robust,o2012unbiased}. Multi-step or multi-level approaches have also been proposed to segment streamlines, for example, by combining both voxel and streamline groupings \cite{guevara2012automatic}, fusing labels from multiple hand-labeled atlases \cite{jin2014automatic}, or using a bundle representation based on maximum density paths \cite{prasad2014automatic}. A few studies have also investigated the representation of specific streamline bundles using different techniques such as gamma mixture models \cite{maddah2008unified}, the computational model of rectifiable currents \cite{durrleman2009statistical,gori2016parsimonious}, and functional varifolds \cite{kumar2017white}. For detailed review of white matter clustering approaches, we refer the reader to \cite{o2013fiber}.
	
	\subsection{Sparse coding for neuroimaging}	
	
	Sparse coding, with an objective of encoding a signal as a sparse combination of prototypes in a data-driven dictionary, has been applied in various domains of computer vision and pattern recognition \cite{elad2010role,wright2009robust,wright2010sparse,yang2009linear}. Various neuroimaging applications have also utilized concepts from this technique, such as the reconstruction \cite{lustig2008compressed} or segmentation \cite{tong2013segmentation} of MRI data, and for functional connectivity analysis \cite{lee2016spark,lee2016sparse}. For diffusion data, sparse coding has been used successfully for clustering white matter voxels from Orientation Density Function (ODF) data \cite{ccetingul2014segmentation}, and for finding a population-level dictionary of key white matter tracts \cite{zhu2016population}.  
	
	Recently, several studies have outlined the connection between clustering and factorization problems, such as dictionary learning \cite{aharon2006svd,sprechmann2010dictionary} and non-negative matrix factorization \cite{kim2007sparse}. Thus, dictionary learning can be seen as a soft clustering, where objects can be linked to more than one cluster. Researchers have also recognized the advantages of applying kernels to existing clustering methods, like the k-means algorithm \cite{dhillon2004kernel}, as well as dictionary learning approaches \cite{nguyen2012kernel}. Such ``kernel'' methods have been shown to better learn the non-linear relations in the data \cite{hofmann2008kernel}.
	
	Sparse coding and dictionary learning were used in \cite{moreno2016sparse,alexandroni2017white} to obtain a compressed representation of streamlines. In our previous work \cite{kumar2015brain,kumar2016sparse}, we applied these concepts to learn an multi-subject streamline atlas for labelling the streamlines of a new subject. In recent studies, we showed how this idea can be used to derive a brain fingerprint capturing genetically-related information on streamline geometry \cite{kumar2017fiberprint}, and to incorporate along-tract measures of micro-structure in the representation \cite{kumar2017white}.    
	
	The present study extends our preliminary work in \cite{kumar2017white,kumar2015brain,kumar2016sparse,kumar2017fiberprint} by providing an in-depth analysis that compares different sparsity priors and evaluates the impact of various parameters. As algorithmic contributions, we present two extensions of the model in \cite{kumar2016sparse}, based on group sparsity and manifold regularization, that provide more meaningful bundles and can incorporate information on streamline geometry, such as the proximity of streamline endpoints, to constrain the clustering process.
	
	
	\section{Kernel dictionary learning for streamline clustering}
	\label{sec:MaterialsAndMethods}

	In this section we propose kernel dictionary learning and sparsity priors based frameworks for white matter fiber analysis. We start with a brief review of dictionary learning and the k-means algorithm, followed by proposed methods based on various sparsity priors, and algorithm complexity analysis.	
	
	\subsection{Dictionary learning and the k-means algorithm}
	\label{sec:DL_and_KM}
	
	Let $\XX$ be the set of $n$ streamlines, each represented as a set of 3D coordinates. For the purpose of explanation, we suppose that each streamline $i$ is encoded as a feature vector $\xx_i \in \Real^d$, and that $\XX$ is a $d \! \times \! n$ feature matrix. Since our dictionary learning method is based on kernels, a fixed set of features is however not required, and streamlines having a different number of 3D coordinates could be compared with a suitable similarity measure (i.e., the kernel function).
	
	The traditional (hard) clustering problem can be defined as assigning each streamline to a bundle from a set of $m$ bundles, such that streamlines are as close as possible to their assigned bundle's prototype (i.e., cluster center). Let $\Psi^{m \times n}$ be the set of all $m \! \times \! n$ cluster assignment matrices (i.e., matrices in which each row has a single non-zero value equal to one), this problem can be expressed as finding the matrix $\DD$ of $m$ bundle prototypes and the streamline-to-bundle assignment matrix $\WW$ that minimize $\|\XX - \DD\WW\|_F^2$. 
	This formulation of the clustering problem can be seen as a special case of dictionary learning, where $\DD$ is the dictionary and $\WW$ is constrained to be a cluster assignment matrix, instead of enforcing its sparsity.
	
	While solving this clustering problem is NP-hard, optimizing $\WW$ or $\DD$ individually is easy. For a given dictionary $\DD$, the optimal $\WW$ assigns each streamline $i$ to the prototype $m$ closest to its feature vector. Likewise, for a fixed $\WW$, the optimal dictionary is found by solving a simple linear regression problem. This simple heuristic correspond to the well-known k-means algorithm.
	
	\subsection{Kernel k-means}
	\label{sec:KKM}
	
	In our streamline clustering problem, the k-means approach described in the previous section has two important disadvantages. First, it requires to encode streamlines as a set of features, which is problematic due to the variation in their length and endpoints. Also, it assumes linear relations between the streamlines and bundle prototypes, while these relations could be better defined in a non-linear subspace (\emph{manifold}).
	
	These problems can be avoided by using a kernel version of k-means for the streamline clustering problem. In this approach, each streamline is projected to a $q$-dimensional space using a mapping function $\pphi : \Real^d \to \Real^q$, where $q \gg d$. We denote by $\PPhi$ the $\Real^{q \times n}$ matrix containing the tracts of $\XX$ mapped with $\pphi$. The inner product of two streamlines in this space corresponds to a kernel function $k$, i.e. $  k(\xx_i, \xx_j) \ = \ \tr{\pphi(\xx_i)}\pphi(\xx_j)$. With $\KK  \ = \ \tr{\PPhi}\PPhi$, the kernel matrix, the \emph{kernel} clustering problem can be expressed as:
	\beq\label{eqn:KKM_cost}
	\argmin_{\substack{\DD \, \in \, \Real^{q \times k} \\ 
			\WW \, \in \, \{0,1\}^{m \times n}}} \
	\|\PPhi - \DD\WW\|_F^2 \quad \tx{subject to} \ \ \tr{\WW}\vo_m \ = \ \vo_n.
	\eeq
	Since the dictionary prototypes are defined in the kernel space, $\DD$ cannot be computed explicitly. To overcome this problem, we follow the strategy proposed in \cite{nguyen2012kernel,rubinstein2010double} and define the dictionary as $\DD = \PPhi \AAA$, where $\AAA \in \Real^{n \times m}$.
	
	Using a similar optimization approach as in k-means, we alternate between updating matrix $\WW$ and $\AAA$. Thus, we update $\WW$ by assigning each streamline $i$ to the prototype $m$ whose features in the kernel space are the closest:
	\beq\label{eqn:W-KKM}
	w_{mi} \ = \ \left\{\begin{array}{ll}
		1 : & \tx{if } m = \arg\min_{m'} \ [\tr{\AAA} \KK \AAA]_{m'm'} \ - \ 2[\tr{\AAA} \kk_i]_{m'}, \\
		0 : & \tx{otherwise}.
	\end{array}\right.,
	\eeq 
	where $\kk_i$ corresponds to the $i$-th column of $\KK$. Recomputing $\AAA$ corresponds once again to solving a linear regression problem with optimal solution:
	\beq\label{eqn:A-KKM}
	\AAA \ = \ \tr{\WW} \big(\WW \tr{\WW}\big)^{-1}.
	\eeq  
	We initialize matrix $\AAA$ as a random selection matrix (i.e., random subset of columns in the identity matrix). This is equivalent to using a random subset of the transformed streamlines (i.e., subset of columns in $\PPhi$) as the initial dictionary. This optimization process is known as \emph{kernel k-means} \cite{dhillon2004kernel}.

	\subsection{Non-negative kernel sparse clustering}
	\label{sec:KSC}
	
	Because they map each streamline to a single bundle, hard clustering approaches like (kernel) k-means can be sensitive to poorly separated bundles and streamlines which do not fit in any bundle (outliers). This section describes a new clustering model that allows one to control the hardness or softness of the clustering. 
	
	In the proposed model, the hard assignment constraints are replaced with non-negativity and $L_0$-norm constraints on the columns of $\WW$. Imposing non-negativity is necessary because the values of $\WW$ represent the membership level of streamlines to bundles. Moreover, since the $L_0$-norm counts the number of non-zero elements, streamlines can be expressed as a combination of a small number of prototypes, instead of a single one. 
	When updating the streamline-to-bundle assignments, the columns $\ww_i$ of $\WW$ can be optimized independently, by solving the following sub-problem: 
	\beq\label{eqn:KSC_cost}
	\argmin_{\ww_i \, \in \, \Real_+^m} \ \|\phi(\xx_i) - \PPhi \AAA \ww_i\|^2_2 \quad 
	\tx{subject to} \ \ \|\ww_i\|_0 \leq S_\mr{max}.
	\eeq
	Parameter $S_\mr{max}$ defines the maximum number of non-zero elements in $\ww_i$ (i.e., the sparsity level), and is provided by the user as input to the clustering method.
	
	The algorithm summary and computational complexity is reported in Supplement material, Algorithm 1. To compute non-negative weights $\ww_i$, we modify the kernel orthogonal matching pursuit (kOMP) approach of \cite{nguyen2012kernel} to include non-negativity constrains of sparse weights (Supplement material, Algorithm 2). Unlike kOMP, the most \emph{positively} correlated atom is selected at each iteration, and the sparse weights $\ww_s$ are obtained by solving a non-negative regression problem. Note that, since the size of $\ww_s$ is bounded by $S_\mr{max}$, computing $\ww_s$ is fast.
	
	In the case of a soft clustering (i.e., when $S_\mr{max} \geq 2$), updating $\AAA$ with (\ref{eqn:A-KKM}) can lead to negative values in the matrix. As a result, the bundle prototypes may lie outside the convex hull of their respective streamlines. To overcome this problem, we adapt a strategy proposed for non-negative tri-factorization \cite{ding2006orthogonal} to our kernel model. In this strategy, $\AAA$ is recomputed by applying the following update scheme, until convergence:
	\beq\label{eqn:A-KSC}
	[\AAA]_{ij} \ \gets \ [\AAA]_{ij} \cdot \dfrac {\left[ \KK \tr{\WW}\right]_{ij}} {\left[\KK \AAA \WW \tr{\WW}\right]_{ij}}, \quad i=1,\ldots,n, \quad j=1,\ldots,m.
	\eeq
	The above update scheme produces small positive values instead of zero entries in $\AAA$. To resolve this problem, we apply a small threshold in post-processing. 
	In terms of computational complexity, the bottleneck of the method lies in computing the kernel matrix. For large datasets, we could reduce this computational complexity by approximating the kernel matrix with the Nystr\"om method \cite{fowlkes2004spectral,Odonnell07automatictractography} (Supplement material, Section 1.5).

	
	\subsection{Extension 1: group sparse kernel dictionary learning}
	\label{sec:GKSC_L1_L21}
	
	The methods proposed above may find insignificant bundles (e.g., bundles containing only a few streamlines) when the parameter controlling the number of clusters is not properly set. Due to the lack of gold standard in tractography analysis, finding a suitable value for this parameter is challenging. 
	
	To overcome this problem, we present a new clustering method based on group sparse kernel dictionary learning. We reformulate the clustering problem as finding the dictionary $\DD$ and non-negative weight matrix $\WW$ minimizing the following problem:  
	\beq\label{eqn:GKSC_cost}
	\argmin_{\substack{\AAA \in \Real^{n \times m} \\ 
			\WW \, \in \, \Real_+^{m \times n}}} \ 
	\frac{1}{2}\|\PPhi - \PPhi\AAA\WW\|_F^2 \, + \, \lambda_1 \|\WW\|_1 \, + \, \lambda_2 \|\WW\|_{2,1}.
	\eeq
	In this formulation, $\|\WW\|_1 = \sum_{i=1}^K\sum_{j=1}^N |w_{ij}|$ is an $L_1$ norm prior which enforces global sparsity of $\WW$, and $\|\WW\|_{2,1} = \sum_{i=1}^K \|\wri\|_2$ is a mixed $L_{2,1}$ norm prior imposing the vector of row norms to be sparse. Concretely, the $L_1$ norm prior limits the ``membership'' of streamlines to a small number of bundles, while the $L_{2,1}$ prior penalizes the clusters containing only a few streamlines. 
	%
	Parameters $\lambda_1, \lambda_2 \geq 0$ control the trade-off between these three properties and the reconstruction error (i.e., the first term of the cost function). 
	
	
	We solve this problem using an Alternating Direction Method of Multipliers (ADMM) algorithm \cite{boyd2011distributed}. First, we introduce ancillary matrix $\ZZ$ and reformulate the problem as: 
	\beq\label{eqn:W-GKSC_L1_L21_cost}
	\argmin_{\substack{\AAA \in \Real_+^{n \times m} \\ \WW, \ZZ \, \in \, \Real_+^{m \times n}}} \
	\frac{1}{2}\|\PPhi - \PPhi \AAA \WW\|^2_F 
	\, + \, \lambda_1 \|\ZZ\|_1 \, + \, \lambda_2 \|\ZZ\|_{2,1} \ \ \tx{subject to} \ \WW = \ZZ.
	\eeq
	We then convert this an unconstrained problem using an Augmented Lagrangian formulation with multipliers $\UU$:
	\beq\label{eqn:W-GKSC_L1_L21_aug_lag}
	\argmin_{\substack{\AAA \in \Real_+^{n \times m} \\ \WW, \ZZ \, \in \, \Real_+^{m \times n}}} \ 
	\frac{1}{2}\|\PPhi - \PPhi \AAA \WW\|^2_F 
	\, + \, \lambda_1 \|\ZZ\|_1 \, + \, \lambda_2 \|\ZZ\|_{2,1} 
	\, + \, \frac{\mu}{2}\|\WW - \ZZ + \UU\|_F^2.
	\eeq  
	Parameters $\WW$, $\ZZ$ and $\UU$ are updated alternatively until convergence. In this work, we use primal feasibility as convergence criteria and stop the optimization once $\|\WW - \ZZ\|_F^2$ is below a small epsilon. 
	
	Dictionary matrix $\AA$ is updated as (\ref{eqn:A-KSC}). To update $\WW$, we derive the objective function with respect to this matrix and set the result to $0$, yielding: 
	\beq\label{eqn:W-GKSC_L1_L21_W}
	\WW \ = \ \big(\tr{\AAA}\KK\AAA + \mu \II\big)^{-1}\big(\tr{\AAA}\KK + \mu(\ZZ-\UU)\big).
	\eeq
	Note that imposing non-negativity on $\WW$ is not required since we ensure this property for $\ZZ$ and have $\WW \approx \ZZ$ at convergence.
	
	Optimizing $\ZZ$ corresponds to solving a group sparse proximal problem \cite{friedman2010note}. This can be done in two steps. First, we do a $L_1$-norm shrinkage by applying the non-negative soft-thresholding operator to each element of $\WW+\UU$:
	\beq\label{eqn:W-GKSC-l1thres}
	\hat{z}_{ij} \ = \ S^+_{\nicefrac{\lambda_1}{\mu}}\big(w_{ij} + u_{ij}\big) \ = \ 
	\max\Big\{w_{ij} + u_{ij} - 
	\nicefrac{\lambda_1}{\mu}, \, 0\Big\}, \quad i \leq K, \ j \leq N.
	\eeq 
	Then, $\ZZ$ is obtained by applying a group shrinkage on each row of $\hat{\ZZ}$:
	\beq\label{eqn:W-GKSC-l2l1thres}
	\zri \ = \ \max\Big\{\|\zhri\|_2 - \nicefrac{\lambda_2}{\mu}, \, 0\Big\} \cdot \frac{\zhri}{\|\zhri\|_2}, \quad i \leq K.
	\eeq
	Finally, the Lagrangian multipliers are updated as in standard ADMM methods: $\UU \ := \ \UU  + (\WW - \ZZ)$.
	The overall optimization procedure and its computational complexity are reported in Supplement material, Algorithm 3.	
	
	
	\subsection{Extension 2: kernel dictionary learning with manifold prior}
	\label{sec:GKSC_L1_Lap}	
	
	Another challenge in streamline clustering is to generate anatomically meaningful groupings. This may require incorporating prior information into the clustering process, for example, to impose streamlines ending in the same anatomical region to be grouped together. In this work, we address this challenge by proposing a manifold-regularized kernel dictionary learning method. 
	
	In the proposed method, we define the manifold as a graph with adjacency matrix $\GG \in \Real^{n \times n}$. In this matrix, $g_{i,i'} = 1$ if streamlines $i$ and $i'$ should be grouped in the same bundle, otherwise $g_{i,i'} = 0$. The manifold regularization prior on the streamline-to-bundle assignments can be formulated as
	\begin{align}\label{eqn:manifold-reg}
		\Reg_\mr{man}(\WW) & \ = \ \lambda_L \! \sum_{i=1}^n \sum_{i'=1}^n g_{i,i'} \, \|\ww_i - \ww_{i'}\|_2^2 \nonumber\\
		& \ = \ \lambda_L \, \mr{tr}(\WW\LLL\tr{\WW}),
	\end{align}
	where $\LL \in \Real^{n \times n}$ is the Laplacian of $\GG$ and $\lambda_L$ is a parameter controlling the strength of constraints on streamlines. 
	
	Our manifold-regularized formulation is obtained by replacing the $L_{2,1}$ prior on $\WW$ with $\Reg_\mr{man}(\WW)$. This new formulation can be solved, as the previous one, with an ADMM algorithm. The main difference occurs when updating $\WW$, which corresponds to the following problem:
	\beq\label{eqn:W-GKSC_Lap_aug_lag}
	\begin{split}
		\argmin_{\WW \, \in \, \Real^{k \times n}} \ 
		\|\PPhi - \PPhi \AAA \WW\|^2_F \, + \, \lambda_L\,\mr{tr}(\WW\LLL\tr{\WW})
		\, + \, \mu\|\WW - \ZZ + \UU\|_F^2 .
	\end{split}
	\eeq  
	Derive this objective function with respect to $\WW$ and setting the result to $0$ gives a Sylvester equation of the form $\PP \WW + \WW \QQ = \RR$ where, $\PP = \tr{\AAA}\KK\AAA + \mu\II$, $\QQ = \lambda_L \LLL$, and $\RR = \tr{\AAA}\KK + \mu(\ZZ-\UU)$. This equation can be solved using Bartels-Stewart algorithm \cite{bartels1972solution}, which requires transforming $\PP$ and $\QQ$ into Schur form with a QR algorithm, and solving the resulting triangular system via back-substitution (Supplement material, Algorithm 4). The computational complexity is $O(n^3)$, $n$ being the size of $\QQ$. However, this can be drastically reduced by pre-computing once the Schur form of $\QQ$.

	
	\section{Experimental results and analysis}
	\label{sec:Results}
	
	In this section, we evaluate our proposed methods on a labeled dataset, followed by parameter impact analysis, and concluding with Human Connectome Project data results on clustering and automated segmentation of new subjects. 
	
	\subsection{Data and pre-processing}
	\label{subsec:DataAndPreProc}
	
	In the first experiment, we compared the proposed methods on a dataset of manually/expert labeled streamline bundles provided by the Sherbrooke Connectivity Imaging Laboratory (SCIL). The source dMRI data was acquired from a 25 year old healthy right-handed volunteer and is described in \cite{fortin2012tractography}. We used 10 of the largest bundles, consisting of 4449 streamlines identified from the cingulum, corticospinal tract, superior cerebellar penduncle and other prominent regions. Figure \ref{fig:Fiber_bundles_GT_vis_compare} (left) shows the coronal and sagittal plane view of the ground truth set. Fibernavigator tool \cite{chamberland2014real} was used for visualizations of this dataset. 
	
	To evaluate the performance of our method across a population of subjects, we two datasets. First, consisting of 12 healthy volunteers (6 males and 6 females, between 19 to 35 years of age) from the freely available MIDAS dataset \cite{bullitt2005vessel}. For streamline tractography, we used the tensor deflection method \cite{lazar2003white} with the following parameters: minimum fractional anisotropy of 0.1, minimum streamline length of 100 mm, threshold for streamline deviation angle of 70 degrees. A mean number of 9124 streamlines was generated for the 12 subjects.
	
	Second, used the pre-processed dMRI data of $10$ unrelated subjects (age 22--35) from the Q3 release of the Human Connectome Project \cite{glasser2013minimal,van2012human,van2013wu}, henceforth referred to as HCP data. All HCP data measure diffusivity along 270 directions distributed equally over 3 shells with b-values of 1000, 2000 and 3000 $\nicefrac{\tx{s}}{\tx{mm}^2}$, and were acquired on a Siemens Skyra 3T scanner with the following parameters: sequence = Spin-echo EPI; repetition time (TR) = 5520 ms; echo time (TE) = 89.5 ms; resolution = $1.25 \times 1.25 \times 1.25$ $\tx{mm}^3$ voxels. Further details can be obtained from HCP Q3 data release manual\footnote{\url{http://www.humanconnectome.org/documentation/Q3/}}.
	
	For signal reconstruction and tractography, we used the freely available DSI Studio toolbox \cite{yeh2010generalized}. All subjects were reconstructed in MNI space using the Q-space diffeomorphic reconstruction (QSDR) \cite{yeh2011ntu} option in DSI Studio. We set output resolution to $1$ mm. For skull stripping, we used the masks provided with pre-processed diffusion HCP data. Other parameters were set to the default DSI Studio values. Deterministic tractography was performed with the Runge-Kutta method of DSI Studio \cite{basser2000vivo,yeh2013deterministic}, using the following parameters: minimum length of $40$ mm, turning angle criteria of $60$ degrees, and trlinear interpolation. The termination criteria was based on the quantitative anisotropy (QA) value, which is determined automatically by DSI Studio. As in the reconstruction step, the other parameters were set to the default DSI Studio values. Using this technique, we obtained a total of $50\,000$ streamlines for each subject. 
	
	As a note, whether the streamlines, generated from tractography, represent the actual white matter pathways remains a topic of debate \cite{jones2013white,thomas2014anatomical}. Streamlines derived from DSI studio are hypothetical curves in space that represent, at best, the major axonal directions suggested by the orientation distribution functions of each voxel, which may contain tens of thousands of actual axonal streamlines.
	
	\subsection{Experimental methodology}
	\label{subsec:methodology}
	
	We tested three distance measures used in the literature for the streamline clustering problem: 1) the Hausdorff distance (Haus) \cite{corouge2004towards,o2005white} which measures the maximum distance between any point on a streamline and its closest point on the other streamline, 2) the mean of closest points distance (MCP) \cite{corouge2004towards} that computes the mean distance between any point on a streamline and its closest point on the other streamline, and 3) the end points distance (EP) \cite{moberts2005evaluation} measuring the mean distance between the endpoints of a streamline and the closest endpoint on the other streamline.
	
	Fiber distances were converted into similarities by applying a radial basis function (RBF) kernel: $k_{i,i'} = \exp\big(\!-\!\gamma \, \mr{dist}_{i,i'}^2\big)$. Parameter $\gamma$ was adjusted separately for each distance measure, using the distribution of values in the corresponding distance matrix. Since the tested distance measures are not all metrics, we applied spectrum shift to make kernels positive semi-definite: $\KK_\mr{psd} = \KK + |\lambda_\mr{min}| \, \II$, where $\lambda_\mr{min}$ is the minimum eigenvalue of $\KK$. This technique only modifies self similarities and is well adapted to clustering \cite{chen2009learning}. 
	
	We initialized $\WW$ using the output of a spectral clustering method \cite{o2005white}, which applies the k-means algorithm on the $10$ first eigenvectors of the normalized Laplacian matrix of $\KK$. To avoid inversion problems when $\WW \tr{\WW}$ is close to singular, we used a small regularization value of 1e-8. Finally, to compare our method with hard clustering approaches, we converted its soft clustering output to a hard clustering by mapping each streamline $i$ to the bundle $j$ for which $w_{ji}$ is maximum. 
	
	We compared our kernel sparse clustering (KSC) approach to four other methods: kernel k-means (KKM) using the same $\KK$ and initial clustering, the spectral clustering (Spect) approach described above, single linkage hierarchical clustering (HSL) \cite{moberts2005evaluation}, and QuickBundles (QB) \cite{garyfallidis2012quickbundles}. The performance of these methods was evaluated using four clustering metrics: the Rand Index (RI) which measures the consistency of the clustering output with respect to the ground truth, the Adjusted Rand Index (ARI) adjusting ARI values by removing the chance agreement, the Normalized Adjusted Rand Index (NARI) that further normalizes the values by considering the cluster sizes, and the Silhouette (SI) measure which does not use the ground truth and measures the ratio between the intra-cluster and inter-cluster distances \cite{rousseeuw1987silhouettes}. While RI, ARI and NARI values range from $0.0$ to $1.0$, SI values are between $-1.0$ and $1.0$. In practice, SI values are generally much lower than $1.0$ due to the intrinsic intra-cluster variance. More information about these metrics can be found in \cite{moberts2005evaluation,siless2013comparison}.

		\begin{table}[ht]
			\caption{Clustering accuracy of our KSC method ($S_\mr{max}\!=\!3$), kernel k-means (KKM), spectral clustering (Spect), and hierarchical clustering (HSL), using the Hausdorff, MCP and EP distances, on the SCIL dataset. For KSC, KKM and Spect, the mean accuracy over $10$ initializations with $m$=10 is reported. The best results for a distance and accuracy metric are shown in boldface type. 
			}
			{
				\centering
				\fontsize{8}{8}\selectfont
				\begin{tabular}{ l l c c c c}
					\toprule
					\multirow{2}{*}{\textbf{Dist}}  & \multirow{2}{*}{\textbf{Method}}  & \textbf{RI}  & \textbf{ARI}  & \textbf{NARI} & \textbf{SI}\\
					&  & mean (std) & mean (std) & mean (std) & mean (std) \\
					\midrule\midrule
					\multirow{5}{*}{MCP}  &  KSC           & \textbf{0.948 (0.012)} & \textbf{0.780 (0.051)} & \textbf{0.716 (0.047)} & \textbf{0.543 (0.032)} \\
					& KKM                 & 0.947 (0.011) & 0.777 (0.049) & \textbf{0.716 (0.046)} & 0.541 (0.028)\\
					& Spect                & 0.942 (0.014) & 0.752 (0.058) & 0.701 (0.047) & 0.515 (0.059)\\
					&  HSL 		& 0.915 (0.000) & 0.704 (0.000) & 0.612 (0.000) & 0.474 (0.000)\\
					& QB            		& 0.943 (0.000) & \textbf{0.780 (0.000)} & 0.696 (0.000) & 0.486 (0.000)\\
					\midrule
					\multirow{5}{*}{Haus}  &  KSC            & \textbf{0.924 (0.013)} & \textbf{0.658 (0.068)} & \textbf{0.634 (0.030)} & \textbf{0.425 (0.022)}\\
					&  KKM                & 0.904 (0.020) & 0.589 (0.082) & 0.573 (0.068) & 0.365 (0.054)\\
					& Spect               & 0.884 (0.018) & 0.517 (0.041) & 0.538 (0.054) & 0.317 (0.069)\\
					& HSL 			& 0.891 (0.000) & 0.640 (0.000) & 0.609 (0.000) & 0.221 (0.000)\\  
					& QB           		& 0.851 (0.000) & 0.468 (0.000) & 0.485 (0.000) & 0.143 (0.000) \\
					\midrule              
					\multirow{5}{*}{EP}  &  KSC                 & \textbf{0.919 (0.005)} & \textbf{0.634 (0.026)} & \textbf{0.641 (0.006)} & \textbf{0.422 (0.020)}\\             
					& KKM                  & 0.915 (0.013) & 0.621 (0.052) & 0.634 (0.034) & 0.410 (0.032)\\                 
					& Spect                 & 0.911 (0.014) & 0.603 (0.053) & 0.616 (0.040) & 0.408 (0.031)\\                     
					& HSL  				& 0.842 (0.000) & 0.539 (0.000) & 0.445 (0.000) & 0.197 (0.000)\\
					& QB             		& 0.885 (0.000) & 0.534 (0.000) & 0.550 (0.000) &  0.129 (0.000)\\    
					
					\bottomrule
				\end{tabular}
			}
			\label{tab:Table_method_distance_comp}
		\end{table}

	\subsection{Comparison of methods and distance measures} 
	
	Table \ref{tab:Table_method_distance_comp} gives the accuracy obtained by KSC ($S_\mr{max}$=3) and the four other tested methods on the SCIL dataset, for the same number of clusters as the ground truth ($m$=10). Since the output of spectral clustering depends on the initialization of its k-means clustering step, for Spect, KSC and KKM, we report the mean performance and standard deviation obtained using $10$ different random seeds. We see that our KSC method improves the initial solution provided by spectral clustering, and gives in most cases a higher accuracy than other clustering methods. We also observe that KSC is more robust to the choice of distance measure than other methods and, as reported in \cite{moberts2005evaluation}, that MCP is consistently better than other distance measures.
	
	\begin{figure}
		{\centering
			\includegraphics{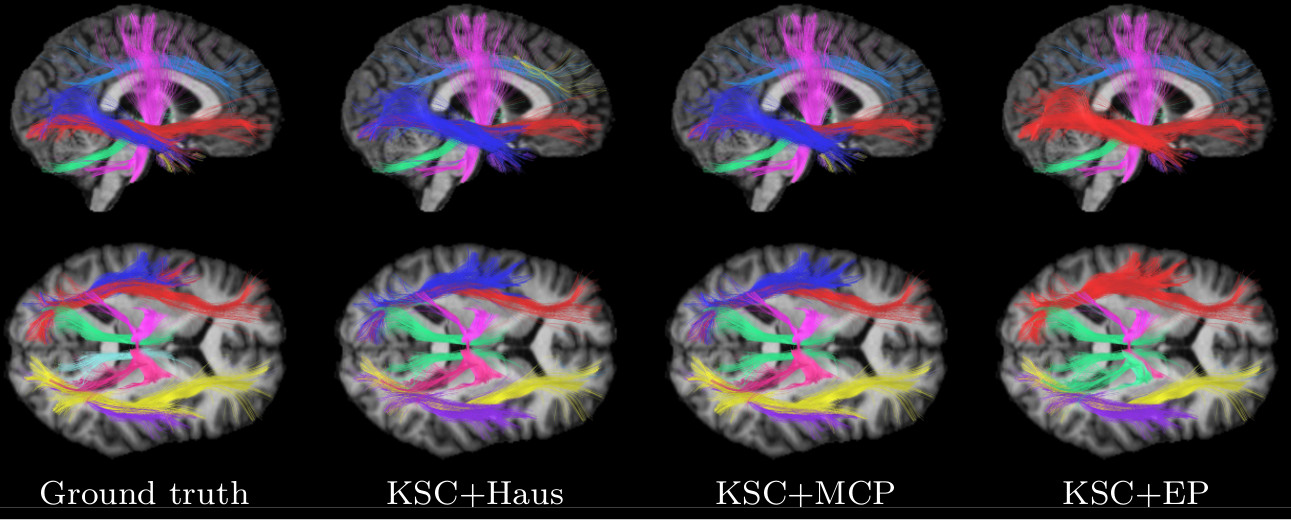}
			
			\vspace{-3mm}
			\caption{Right sagittal (\textbf{top}) and inferior axial (\textbf{bottom}) views of the ground truth, and bundles obtained by KSC ($S_\mr{max}=3$) using the Haus, MCP and EP.}
			\label{fig:Fiber_bundles_GT_vis_compare}
		}
	\end{figure}

	Figure \ref{fig:Fiber_bundles_GT_vis_compare} compares the ground truth clustering of the SCIL dataset with the outputs of KSC ($S_\mr{max}$=3) using the Haus, MCP and EP distances. Except for the superior cerebellar peduncle bundle (cyan and green colors in the ground truth), the bundles obtained by KSC+MCP and KSC+Haus are similar to those of the ground truth clustering. Also, we observe that the differences between KSC+MCP and KSC+Haus occur mostly in the right inferior fronto-occipital fasciculus and inferior longitudinal fasciculus  bundles (yellow and purple colors in the ground truth). Possibly due to the large variance of endpoint distances in individual bundles, KSC+EP gives poor clustering results.
		
	\subsection{Impact of sparsity} 
	
\begin{figure}[t]
		{\centering    
			\mbox{
				\includegraphics[width=11cm]{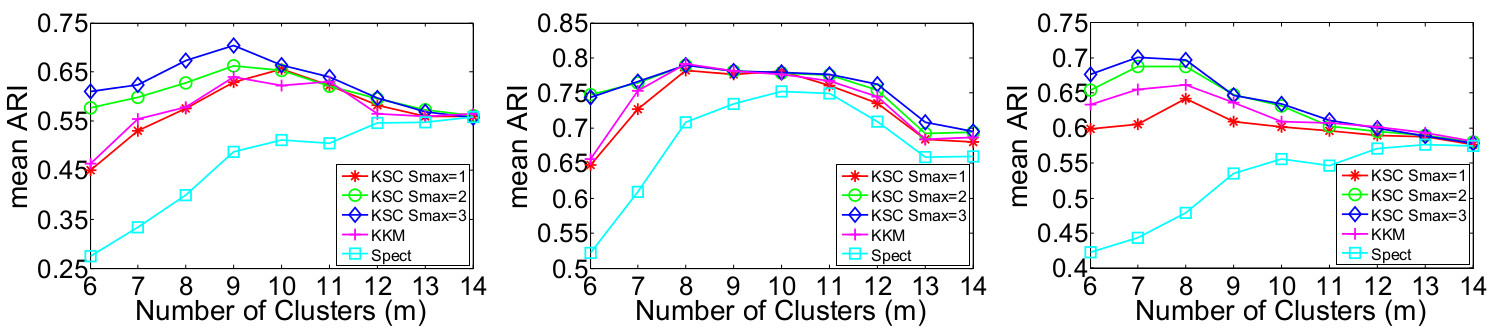}
			}
			
			\vspace{-3mm}
			\caption{Mean ARI obtained on the SCIL dataset by KSC ($S_\mr{max} = 1,2,3$), KKM and Spect, using Haus (\textbf{left}), MCP (\textbf{center}), EP  (\textbf{right}); for varying $m$.}
			\label{fig:GT_KSC_KKM_Spect_comp_MCP}
		}
	\end{figure}

	Figure \ref{fig:GT_KSC_KKM_Spect_comp_MCP} reports the mean ARI (over $10$ runs) obtained on the SCIL dataset by our KSC approach, using $S_\mr{max}$=1,2,3, for an increasing number of clusters (i.e., dictionary size $m$). For comparison, the performance of KKM and Spect is also shown. When the Spectral Clustering initialization is near optimal (i.e., when $m$ is near the true number of clusters and using MCP), both methods find similar solutions. However, when the initial spectral clustering is poor (e.g., Haus and EP distance or small number of clusters) the improvement obtained by KSC is more significant than KKM. Hence, KSC ($S_\mr{max}\!\geq\!2$) is more robust than hard clustering approaches (i.e., Spect, KKM or KSC with $S_\mr{max}$=1) to the number of clusters and distance measures. 
	
		\begin{figure}[h]
			{\centering
				
				\includegraphics[width=11cm]{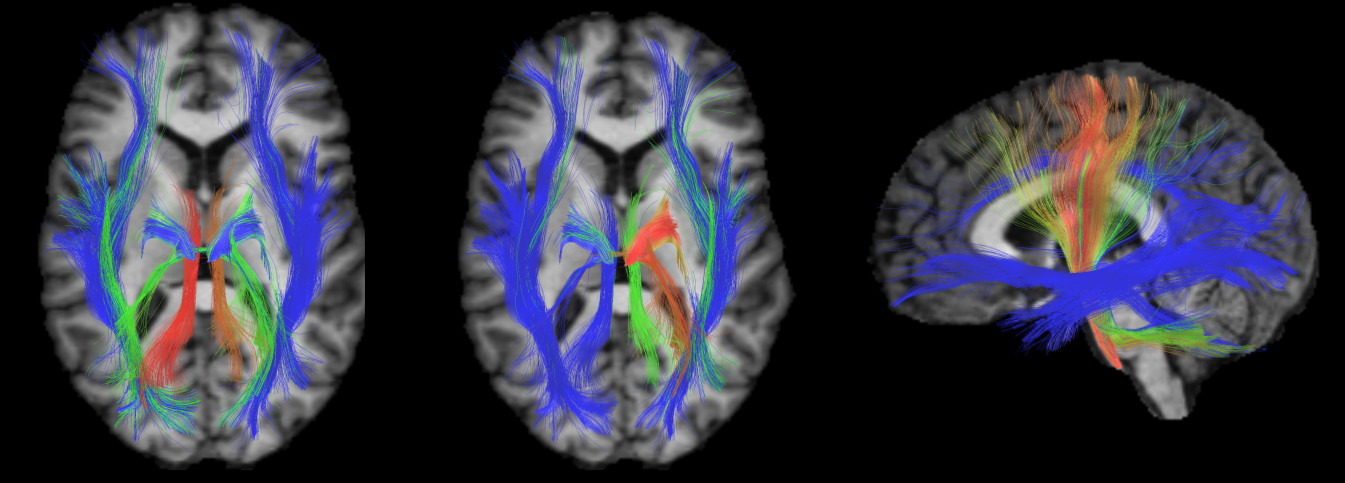}
				
			}
			\vspace{-3mm}
			\caption{Membership level of streamlines to two different bundles (\textbf{left} and \textbf{center}), and importance of each streamline in defining the prototype of a bundle (\textbf{right}). Blue means a null membership/importance, while non-zero values are represented by a color ranging from green (lowest value) to red (highest value).}
			\label{fig:A_W_sparsity_KSC_MCP}
		\end{figure}
	
	To illustrate the soft clustering of KSC, Fig. \ref{fig:A_W_sparsity_KSC_MCP} (\textbf{left}) and (\textbf{center}) show the membership level of streamlines to two different bundles. Streamline colors in each figure correspond to the values of a row in $\WW$ normalized so that the minimum is 0 (blue) and the maximum is $1$ (red). We observe streamlines having a membership to both bundles (e.g., orange-colored streamlines in the left image), reflecting the uncertainty of this part of the clustering. In Fig. \ref{fig:A_W_sparsity_KSC_MCP} (\textbf{right}), we show the importance of each streamline in defining the prototype of a bundle, using the normalized value of a column in $\AAA$ as colors. It can be seen that only a few streamlines are used to define this bundle, confirming the sparsity of $\AAA$.

	\subsection{Group sparsity prior}

		\begin{figure}[h]
			\centering
			\begin{small}
				
				\includegraphics[width=11cm]{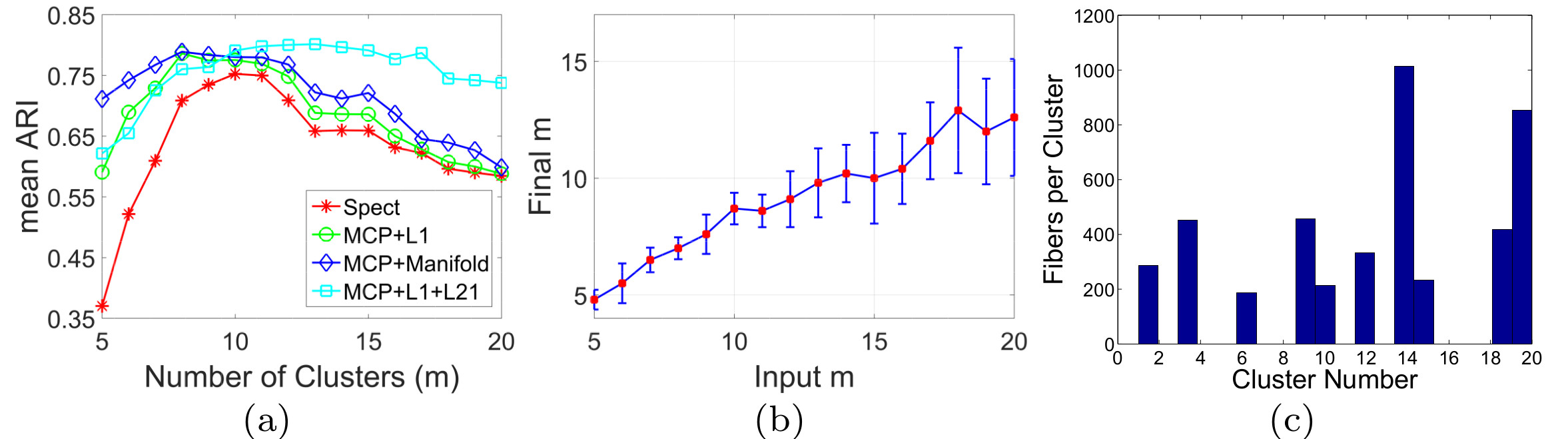}
				\vspace{-3mm}
				\caption{\textbf{(a)} Mean ARI obtained on the SCIL dataset by GKSC, MCP+L1, MCP+Manifold and Spect, using MCP; for varying $m$; \textbf{(b)} mean and standard deviation of final $m$ for varying input $m$; \textbf{(c)} Distribution of bundle sizes for a sample run using $m=20$.}
				\label{fig:GKSC_analysis_final_m}	
			\end{small}
		\end{figure}

	\begin{table}
		\caption{Clustering accuracy of proposed methods
			using MCP distances and three types of priors: $L_1$ norm sparsity alone (L1), with group sparsity (L1+L21), and with manifold regularization (L1+Manifold). Reported values are the mean accuracy over $10$ initializations with (input) $m$=10 clusters. The best result for each accuracy metric is shown in boldface type. 
		}
		{
			\centering
			\fontsize{8.5}{8.5}\selectfont
			\begin{tabular}{l c c c c}
				\toprule
				\multirow{2}{*}{\textbf{Prior}}  & \textbf{RI}  & \textbf{ARI}  & \textbf{NARI} & \textbf{SI}\\
				& mean (std) & mean (std) & mean (std) & mean (std) \\
				\midrule\midrule
				L1          & 0.947 (0.011) & 0.775 (0.049) & 0.714 (0.045) & 0.543 (0.029)\\
				L1+Manifold    & 0.948 (0.010) & 0.780 (0.044) & 0.717 (0.046) & 0.546 (0.033)\\ 
				L1+L21  & \textbf{0.949 (0.006)} & \textbf{0.791 (0.025)} & \textbf{0.721 (0.035)} & \textbf{0.563 (0.039)}\\ 
				\bottomrule
			\end{tabular}
		}
		\label{tab:Table_method_distance_comp_2}
	\end{table}	
	
	Figure \ref{fig:GKSC_analysis_final_m}(a) plots the mean Adjusted Rand Index (ARI) obtained by our group sparse model (MCP+L1+L21) for various cluster numbers ($m$), over $10$ runs with different spectral clustering initializations. As baseline, we also report the ARI of spectral clustering and our method without group sparsity (MCP+L1), i.e. using $\lambda_2$=0. We see that employing group sparsity improves clustering quality and provides a greater robustness to the input value of $m$. The advantages of using a group sparse prior are further confirmed in Table \ref{tab:Table_method_distance_comp_2}, which gives the mean ARI, RI, NARI and average SI for $m$=10. Results show that MCP+L1+L21 outperforms MCP+L1 for all performance metrics. In a t-test, these improvements are statistically significant with $p<$0.01.   
	
	As described in Section \ref{sec:GKSC_L1_L21}, group sparsity has the benefit of providing meaningful bundles, regardless of the number of clusters $m$ given as input. In Fig. 5(a), we see that the ARI of MCP+L1+L21 increases monotonically until reaching the ground-truth number of bundles $m^*$=10. While the clustering accuracy of other methods drops for $m>$10, the performance of MCP+L1+L21 remains stable. This is explained in Fig. (b) which plots the number of non-empty clusters found by MCP+L1+L21 as a function of $m$: the number of output clusters stays near to $m^*$=10, even for large values of $m$. As additional confirmation, Fig. \ref{fig:GKSC_analysis_final_m}(c) shows the number of streamlines per cluster for a sample run of MCP+L1+L21 with $m=20$. In this example, the output clustering contains $m^*$=10 non-empty clusters.

\begin{figure}[h]
		\centering
		\begin{small}
			
			\includegraphics[width=11cm]{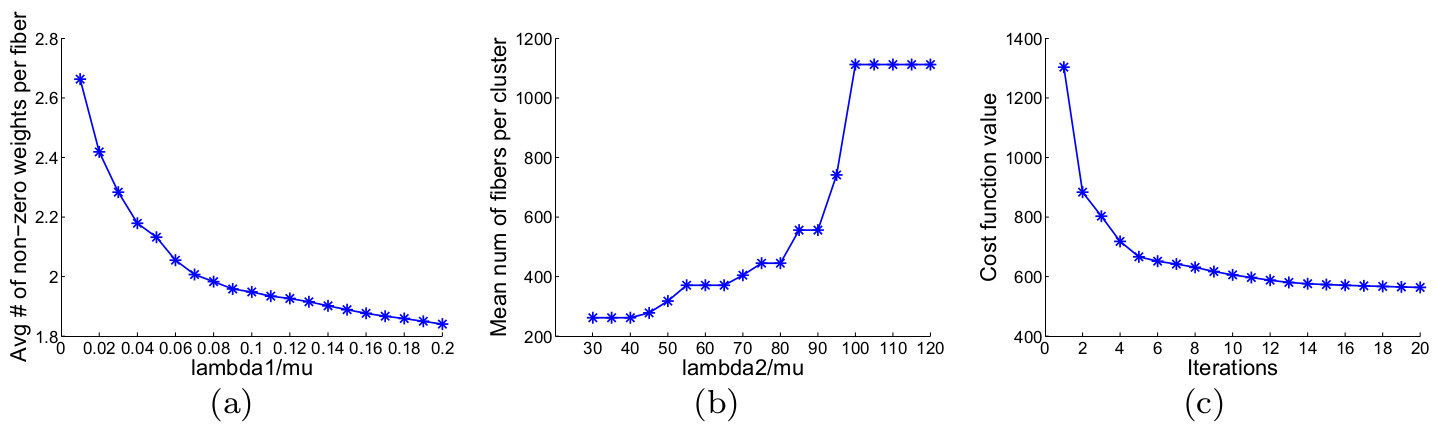}
			\vspace{-3mm}
			\caption{\textbf{(a)} Mean number of non-zero assignment weights per streamline, for $\lambda_2/\mu = 80$ and increasing $\lambda_1/\mu$. \textbf{(b)} Mean number of streamlines per bundle, for $\lambda_1/\mu=0.1$ and increasing $\lambda_2/\mu$. \textbf{(c)} Cost function value at each of a sample run for MCP+L1+L21.}
			\label{fig:Param_var_GKSC}
		\end{small}
	\end{figure}

	In Fig. \ref{fig:Param_var_GKSC}, we measure the impact of sparse regularization parameters $\lambda_1$ and $\lambda_2$ for a fixed ADMM parameter of $\mu=0.01$. As shown in (a), $\lambda_1/\mu$ controls the mean number of non-zero weights per streamline (i.e., how soft or hard is the clustering). Likewise, as illustrated in (b), $\lambda_2/\mu$ defines the size of bundles in the output. These results are consistent with the use $L_1$-norm and $L_{2,1}$-norm sparsity in (\ref{eqn:W-GKSC-l1thres}). Finally, the optimization stability of the MCP+L1+L21 model is illustrated in Fig. \ref{fig:Param_var_GKSC}(c), where convergence is reached around 20 iterations.
	
	\subsection{Manifold regularization prior}
	
	We apply the proposed manifold regularization prior to enforce the grouping of streamlines with similar end-points. The idea is to obtain bundles that correspond to localized regions of the cortex. To generate the Laplacian matrix in (\ref{eqn:manifold-reg}), we constructed a graph where the nodes are streamlines and two nodes are connected if the distance between their nearest endpoints is below some threshold. Following \cite{gori2016parsimonious}, we used a distance threshold of 7mm, giving a Laplacian matrix with overall sparsity near 15\%. 
	
	\begin{figure}[h]
		\centering
		\begin{small}
			
			\includegraphics[width=11cm]{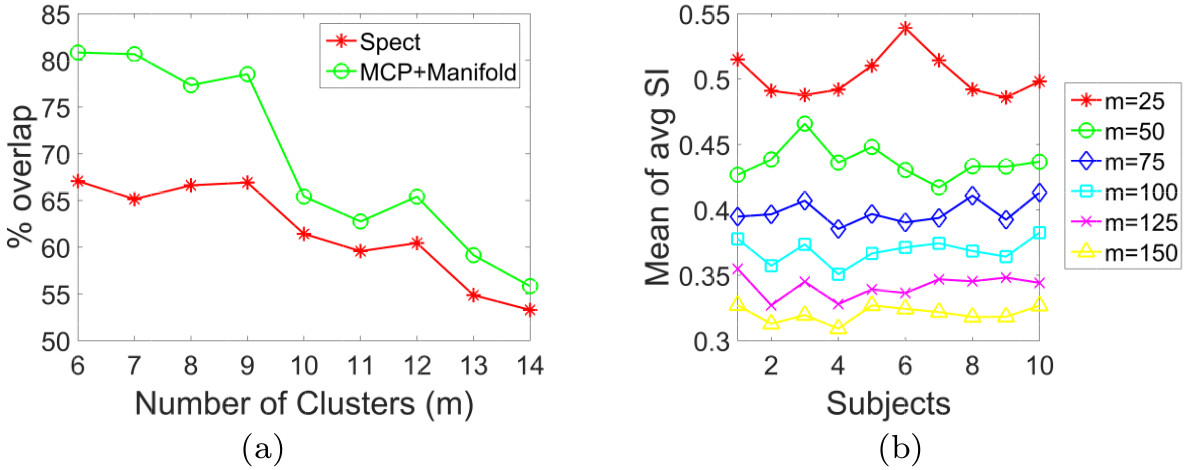}
			\vspace{-4mm}
			\caption{\textbf{(a)} Percentage overlap with EP based Laplacian prior matrix, compared with baseline initialization of spect, for varying $m$. \textbf{(b)} Mean of avg SI for KSC+MCP clustering of $10$ unrelated HCP subjects for varying $m$.}
			\label{fig:Lap_prior_and_HCP_10sub}	
		\end{small}
	\end{figure}

	In Fig. \ref{fig:GKSC_analysis_final_m}(a), we see that the manifold regularization prior (MCP+Manifold) improves performance compared to spectral clustering baseline and $L_1$ norm sparsity (MCP+L1). This improvement is particularly important when the input number of clusters is below that of the ground truth (i.e., $m<$10). Conversely, for $m\!>\!10$, MCP+Manifold is outperformed by group sparsity (MCP+L1+L21) due to the over-segmentation of streamlines. Fig. \ref{fig:Lap_prior_and_HCP_10sub}(a) measures the the percentage of streamlines with nearby endpoints (i.e., edges in the graph) that are assigned to the same cluster, denoted as overlap in the figure. As expected, the prior helps preserve anatomical information defined by streamline endpoints in the clustering.      
	
	
	\subsection{Validation on HCP data}
	
	We evaluated the performance of our kernel sparse clustering (KSC) method on a population of subjects from the Human Connectome Project (HCP). For this experiment, we used two datasets: $10$ unrelated HCP subjects, and subjects from the freely available MIDAS dataset \cite{bullitt2005vessel} (results in Supplement material). The objective here is to show applicability of our method across population-subjects, and analyse the impact of inter-subject variability.  
	
	Figure \ref{fig:Lap_prior_and_HCP_10sub}(b) shows the mean of average SI obtained for the $10$ unrelated subjects, using a varying number $m$ of clusters and $3$ runs for each $m$ value. This plot was generated by sampling $5000$ streamlines uniformly over the full tractography (\cite{o2007automatic,kumar2017fiberprint}) and computing their pairwise MCP distance. We observe that clustering quality decreases with higher values of $m$, and that this quality varies across subjects. A similar trend is observed for MIDAS dataset (Supplement material, Fig. 2). Comparing HCP and MIDAS datasets, a greater average SI is obtained for HCP possibly due to the higher resolution of images in this dataset. Full clustering visualization for $10$ subjects ($m=50$) and subject 1 for $m=25,50,75,100,125,150$  are shown in Supplement material, Figure 3,4). Note the optimal number of streamline clusters is still an open challenge \cite{o2007automatic}, we used $m=50$ in this study for ease of visualization and interpretation. 
	
		\begin{figure}[htb!]
			{\centering 
				
				\includegraphics[scale=0.90]{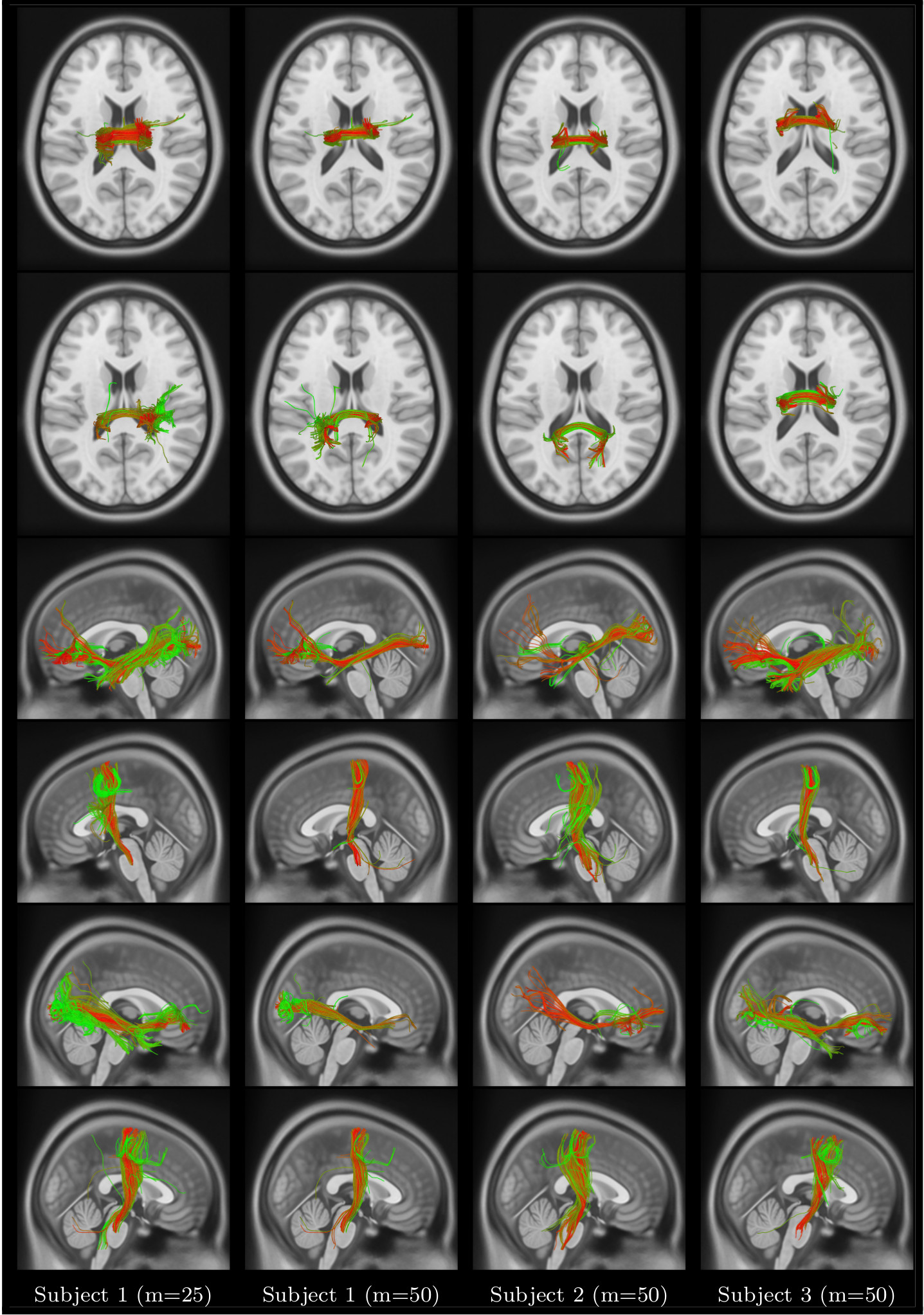}
				
				\vspace{-3mm}
				\caption{Color coded visualization of sparse code memberships of streamlines in Corpus Callosum (row-1,2); left Inferior Occipitofrontal Fasciculus (IOF) and Cortico-Spinal-Tract (CST) (row-3,4); and right IOF and CST (row-5,6).}
				\label{fig:Wsoft_HCP10sub}
			}
		\end{figure}
	
	Figure \ref{fig:Wsoft_HCP10sub} shows sparse code memberships of streamlines in six different bundles: Corpus Callosum - anterior body (row 1) and central body (row 2), left Inferior Occipitofrontal Fasciculus (IOF) (row 3), left Cortico-Spinal-Tract (CST) (row 4), right IOF (row 5), and right CST (row 6). Results are reported for subject 1 ($m$=25 and $m$=50), subjects $2$ ($m$=50) and subject $3$ ($m$=50). Sparse code values are represented by a color ranging from green (lowest value) to red (highest value). While variations are observed across values of $m$ and subjects, the general shape of bundles recovered by our method is similar.
	
		\begin{figure}[htb!]
			{\centering 
				
				\includegraphics[scale=0.90]{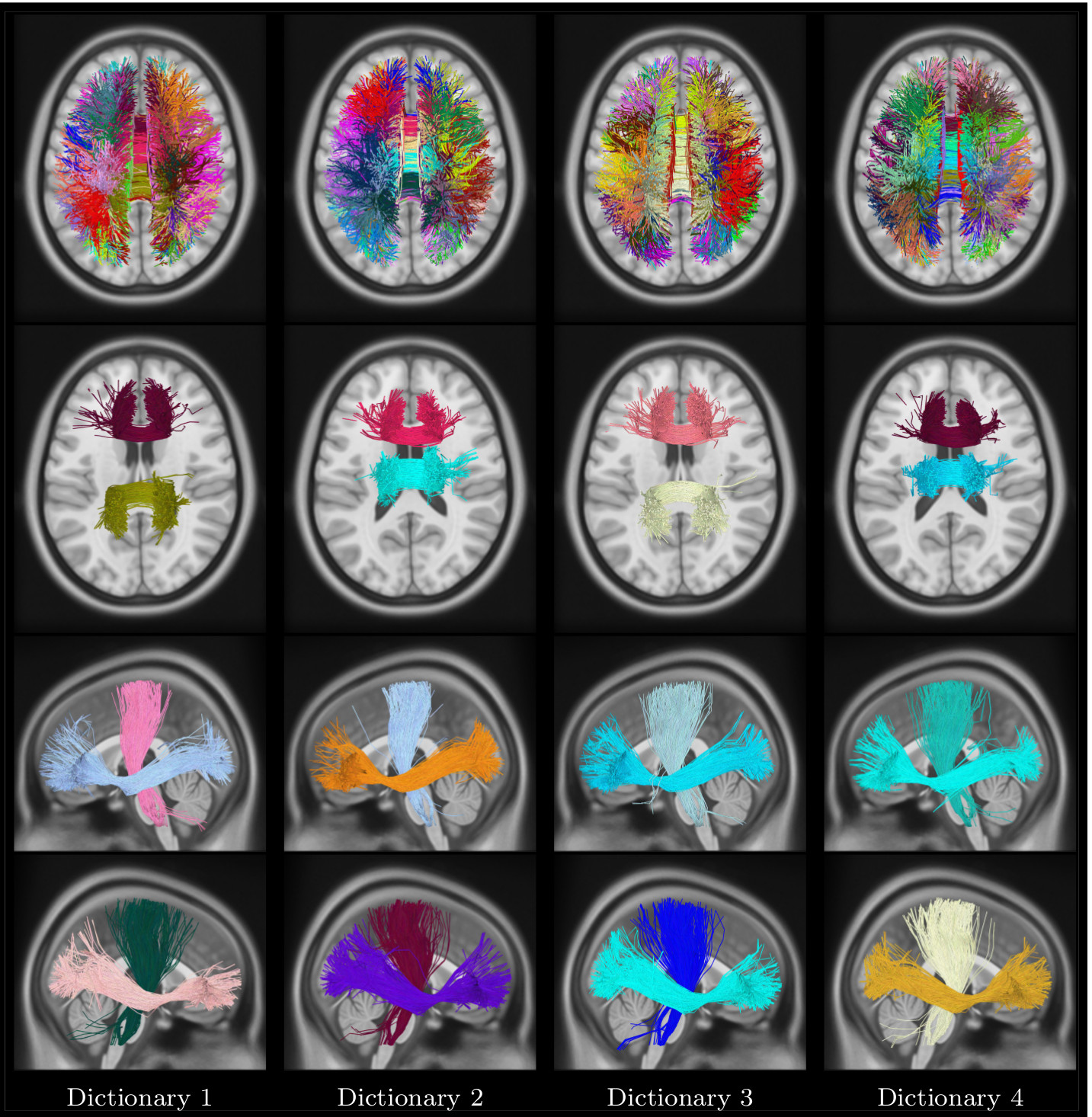}
				
				\vspace{-3mm}
				\caption{Unsupervised multi-subject dictionary visualization. Four different dictionaries and corresponding bundles. Top row: Axial view of full dictionary with a unique color assigned to each bundle; Second row: Anterior Body, and Central Body bundles in Corpus Callosum; Third row: Left CST, and Left IOF bundles; Last row: Right CST, and Right IOF bundles. Each dictionary has a different color code, while the bundles respect that dictionary color-code. (m=50 bundles).}
				\label{fig:Multi-sub-dictionary-sets}
			}
		\end{figure}
		
	\subsection{Application to automated tractography segmentation}
	
	In this section, we apply the proposed KSC method for the automated segmentation of new subject streamlines. Again, the focus of our analysis is on inter-subject variability and its effect on results. To label streamlines, we used as bundle atlas the dictionaries obtained from 40 unrelated HCP subjects ($4$ dictionaries, each one learned from $10$ subjects. Dictionaries were generated by sampling $5000$ streamlines in each subject and employing MCP as distance measure. Note that expert-labeled streamlines could also be used as dictionary. 
	
	\begin{figure}[htb!]
		{\centering 
			
			\includegraphics[scale=0.70]{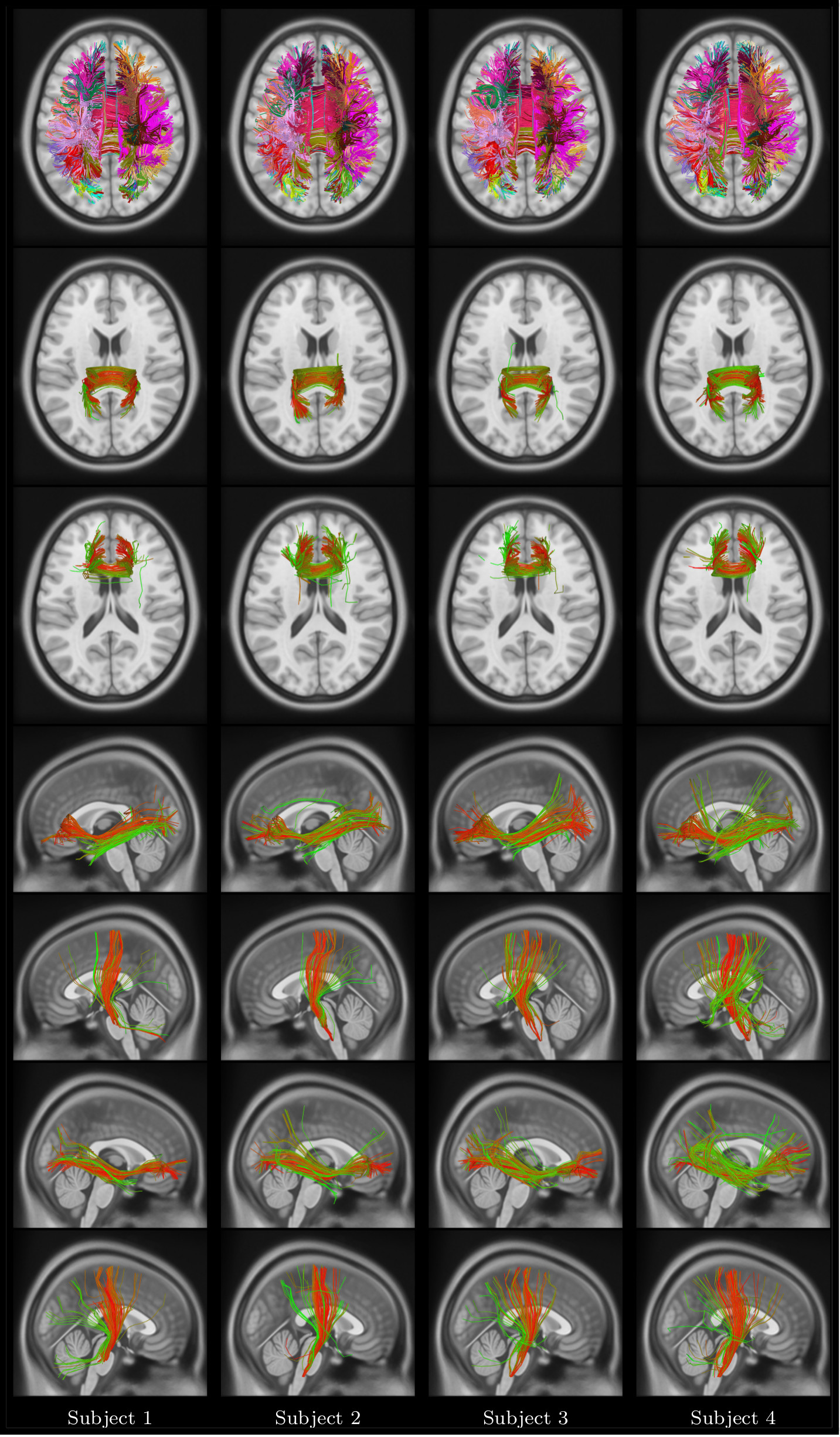}
			
			\vspace{-3mm}
			\caption{Automated segmentation visualization. Top row: full segmentation of 4 HCP subjects using dictionary D1, with a unique color assigned to each cluster, and same color code as D1. Rows 2-7: sparse code (bundle membership) visualization for the posterior body CC, anterior body CC, left IOF, left CST, right IOF, and right CST bundles. Membership values are represented by a color ranging from green (no membership) to red (highest membership).}
			\label{fig:Auto_Seg_4subs_D4}
		}
	\end{figure}

	The bundles encoded by these dictionaries are depicted in Figure \ref{fig:Multi-sub-dictionary-sets}. Moreover, segmentation results obtained for 4 different subjects using dictionary D1 are shown in Fig. \ref{fig:Auto_Seg_4subs_D4}. For each subject, we give the full segmentation as well as membership values for CC, left/right IOF, and left/right CST bundles. Additionally, to analyze the impact of sampling streamlines from a subject, segmentation results for $5$ instances of subject 1 using D1 are provided in Supplement material. Once more, while we observe variability across segmented streamlines from different subjects, the results obtained by our method are globally consistent across subjects. Similar consistency is found across multiple instances of the subject 1 (see Supplement material, Fig. 5). 
		
	\section{Discussion}
	\label{sec:Discussion}
	
	We now summarize and discuss the findings related to proposed approaches, impact of various priors, and their applications. We then highlight limitations and additional considerations of this study.
	
	\subsection{Main findings}
	
	Our experiments have demonstrated the usefulness of our kernel sparse clustering (KSC) and various sparsity priors. The soft assignment provided by KSC ($S_\mr{max}\!\geq\!2$) improved performance for all measures of clustering quality compared to a hard clustering approaches like kernel k-means. This improvement was most significant when the input number of clusters (parameter $m$) is not set close to the ground truth value. In such cases, soft assignment offers a greater robustness to the ambiguous membership of streamlines to bundles. 
	
	Comparing the different streamline distances, we found that mean of closest points (MCP) performed the best. Hausdorff distance measures the maximum distance between any point on a streamline and its closest point on another streamline, and thus fails to capture bundles with branching or diverging streamlines. Likewise, end points distances may be more affected by outlier streamlines or issues in diffusion tractography output. These observations are in line with previous analyses on streamline distances \cite{moberts2005evaluation,siless2013comparison}.
	
	Results revealed the input number of clusters to have a high impact on results. The true value of this parameter is largely unknown \cite{o2007automatic}, and even in expert labeled set could be off the mark due to labeling errors \cite{moberts2005evaluation}. Our analysis showed that group sparsity provides robustness to this confound, and recovers meaningful bundles when it is set far from the ground-truth value. Likewise, the proposed manifold regularization prior helped the clustering by enforcing related pairs of streamlines to be grouped together. This could be useful in a wide range of applications where anatomical information (e.g., cortical parcellation atlas) is available.  
	
	Unsupervised clustering of subjects from HCP and MIDAS datasets showed that our KSC method can be employed for data driven analyses, our method finding plausible clusters corresponding to well known bundles. Moreover, the visualization of clusters and membership values demonstrates that KSC can effectively capture inter-subject variability. Experiments on automated streamline segmentation also revealed that KSC can accurately recover major bundles in new subjects, and that this segmentation is robust to the number of clusters, inter-individual variations, and the sampling of streamlines from the same subject. 
	
	\subsection{Limitations and additional considerations}
	
	Due to the lack of gold standard clustering, as well as the various challenges in diffusion tractography \cite{maier2017challenge} and its interpretation \cite{jones2013white}, validating streamline clustering approaches is difficult. A large scale and data-driven analysis, for example using data from over $1000$ HCP subjects, could lead to interesting observations on number of bundles and their population-wise variability. 
	
	An important aspect of our dictionary learning method is its initialization. While we employed spectral clustering for this task, considering other techniques could possibly lead to better results. For the automated segmentation of streamlines in new subjects, we learned the dictionary in an unsupervised setting, however expert-labeled streamlines set or atlas/clustering from other approaches can also be utilized. 
	
	One the main advantages of the proposed kernel-based framework is that it alleviates the need for an explicit streamline representation. Previous attempts in utilizing dictionary learning and sparse coding for streamline clustering might have been hindered by this. Employing kernels also provides flexibility and enables the extension to other streamline similarity measures, which can incorporate a richer set of characteristics such as along-tract diffusivity \cite{kumar2017white,charon2013varifold,charlier2014fshape}.
	
	Another key element of our study is the anatomical interpretation of clustering results. The streamlines generated from diffusion tractography provide a macro-scale inference of the underlying fibers\cite{jones2013white,maier2017challenge}. As such, the clustering for a given distance/similarity measures focuses primarily on the geometric aspect of streamlines. Although we considered end points proximity in our manifold regularization prior, additional information such as structural parcellation could be incorporated to improve the anatomic plausibility of the final clustering \cite{o2013fiber,siless2018anatomicuts}. 
	
	The sparse code representation of streamlines conveys a wealth of information on inter-individual variability in terms of streamline geometry. Extension of this study could leverage this information for additional tasks, such as identifying noisy/spurious streamlines, discovering tract-based biomarkers to discriminate between healthy and diseased subjects \cite{o2017automated}, or establishing bundle-to-bundle correspondences across subjects.  
	
	
	\section{Conclusion}
	\label{sec:Conclusion}
	
	We presented a novel framework using kernel dictionary learning with various sparsity priors for the unsupervised segmentation of white matter streamlines. The proposed framework does not require explicit streamline representation and enables using any streamline similarity measure. Dictionary bundles are encoded as a non-negative combination of training streamlines, and the kernel trick is used to model non-linear relationships between streamlines and bundles. 
	
	We compared our method against state-of-the-art streamline clustering approaches using expert-labeled data, as well as subjects from the HCP and MIDAS dataset. Results demonstrate the usefulness of having a soft assignment, and that our method is suitable for scenarios where streamlines are not clearly separated, bundles overlap, or when there is important inter-individual variability. Experiments using group sparsity ($L_{2,1}$ norm) and manifold regularization show that these priors can improve clustering quality by adding robustness to the input number of clustering or incorporating anatomical constraints in the clustering. 
	
	The benefits of the proposed approach in cases of inter-individual variability was showcased for the automated segmentation of streamlines from new subjects. In future work, we will investigate the usefulness of our approach for identifying and comparing major bundles in healthy vs diseased subjects, and for incorporating along-tract measures in the clustering process.

	
	\section*{Acknowledgements}
	
	Data were provided in part by the Human Connectome Project, WU-Minn Consortium (Principal Investigators: David Van Essen and Kamil Ugurbil; 1U54MH091657) funded by the 16 NIH Institutes and Centers that support the NIH Blueprint for Neuroscience Research; and by the McDonnell Center for Systems Neuroscience at Washington University. We thank the Sherbrooke Connectivity Imaging Laboratory for generously providing the labeled dataset used in this work.

	
	
	\bibliography{References_KD_Sparsity_journal}

\end{document}


\maketitle
	
	\section{Algorithms}
	\label{sec:SupAlgo}

	\subsection{Non-negative kernelized sparse clustering}
	
	\subsubsection{Algorithm summary}
	\begin{center}
		\begin{small}
			\begin{algorithm2e}[H]
				\KwIn{Pairwise streamline distance matrix $S_{dist} \in \RR^{n \times n}$;}
				\KwIn{The desired number of streamline bundles $m$;}
				\KwIn{The RBF kernel parameter $\gamma$;}
				\KwIn{The sparsity level $S_\mr{max}$ and maximum number of iterations $T_\mr{max}$;}
				\KwOut{The sparse assignment matrix $\WW \in \RR^{n \times m}$ and 
					hard assignment vector $\cc \in \{1,\ldots,m\}^n$;}
				\caption{Algorithm 2: Kernelized sparse clustering method}
				
				\BlankLine
				Initialize the kernel matrix:  $k_{ij} \ = \ \exp(-\gamma \! \cdot \! dist_{ij}^2)$ \;
				Initialize $\AAA$ as a random selection matrix\;
				
				\BlankLine
				\For{$t = 1, \ldots, T_\mr{max}$}{
					Update each column $\ww_i$ of $\WW$ using NNKOMP (Algorithm 2)\;
					Update dictionary until convergence: $ \AAA_{ij} \ \gets \ \AAA_{ij}\dfrac {\left( \KK \tr{\WW}\right) _{ij}} {\left( \KK \AAA \WW \tr{\WW}\right) _{ij}}, \quad i=1,\ldots,n, \quad j=1,\ldots,m.$\;
					$t_\mr{out} \ \gets \ t_\mr{out} + 1$\;
				}	
				
				\BlankLine
				Compute hard assignment: $c_i \ = \	\arg\max _{k'} w_{im'}, \ \ i = 1,\ldots,n$ \;
				
				\BlankLine
				\Return{$\{\WW,c\}$} \;
			\end{algorithm2e}
		\end{small}
	\end{center}
	
	\subsubsection{Algorithm complexity}
	
	In Algorithm 2, the user provides a matrix $S_{dist}$ of pairwise streamline distances, as well as the desired number of bundles (clusters), and obtains in return the soft (matrix $\WW$) and hard (vector $\cc$) streamline clusterings. Various distance measures, suitable for streamlines, are described in experiments. The distances are converted into similarities by using a Gaussian (RBF) kernel of parameter $\gamma$. Note that the obtained kernel is 
	semi-definite positive only if the distance is a metric. However, non-metric distances, such as the Hausdorff distance (see experiments), have been shown to be quite useful in practice \cite{laub2004feature}. In the main loop, the dictionary matrix $\AAA$ and sparse streamline-to-bundle assignment matrix $\WW$ are optimized alternatively, until convergence or $T_\mr{max}$ iterations have been reached. The soft clustering of $\WW$ is converted to a hard clustering by assigning each streamline $i$ to the bundle $m$ for which $w_{im}$ is maximum.
	
	\subsection{Non-negative kernelized orthogonal matching pursuit (NNKOMP)}
	{
		\fontsize{7}{7}\selectfont
		\begin{algorithm2e}[htb!]
			\KwIn{The dictionary matrix $\AAA \in \RR_+^{n \times m}$ and kernel matrix $\KK \in \RR^{n \times n}$;}
			\KwIn{The streamline index $i$ and sparsity level $S_\mr{max}$;}
			\KwOut{The set of non-zero weights $I_s$ and corresponding weight values $\ww_s$;}
			\caption{Non-negative kernelized orthogonal matching pursuit}	
			
			\BlankLine
			Initialize set of selected atoms and weights: $I_0 \ = \ \emptyset, \ \ww_0 \ = \ \emptyset$\;
			
			\BlankLine		
			\For{$s = 1, \ldots, S_\mr{max}$}{
				
				\BlankLine	
				$\tau_j \ = \ \big[\tr{\AAA}\big(\kk_i - \KK \, \AAA_{[I_s]} \ww_s\big)\big]_j \, / \,
				\big[\tr{\AAA}\KK\AAA\big]_{jj}, \ \ j = 1, \ldots, m$\;
				$ j_\mr{max} \ = \ \argmax_{j \, \not \in \, I_{s-1}} \ \tau_j, \ \
				I_s \ = \ I_{s-1} \, \cup \, j_\mr{max}$\;
				$\ww_s \ = \ \argmin_{\ww \, \in \, \RR_+^s} \ 
				\tr{\ww}\tr{\AAA}_{[I_s]} \KK \AAA_{[I_s]}\ww \ - \ 2\, \tr{\kk_i}\AAA_{[I_s]}\ww$\;
			}
			
			\BlankLine
			\Return{$\{I_s, \, \ww_s\}$} \;
			
			\BlankLine
			\textbf{Note:} $\AAA_{[I_s]}$ contains the columns of $\AAA$ whose index is in $I_s$ \;
		\end{algorithm2e}
	}
	
	\subsection{Group sparse kernelized dictionary learning}
	
	\subsubsection{Algorithm summary}
	
	\begin{small}
		\centering
		\begin{algorithm2e}[!htb]
			\KwIn{Pairwise streamline distance matrix $S_{dist} \in \RR^{n \times n}$;}
			\KwIn{The maximum number of streamline bundles $m$;}
			\KwIn{The RBF kernel parameter $\gamma$;}
			\KwIn{The cost trade-off parameters $\lambda_1, \lambda_2$ and Lagrangian parameter $\mu$;}
			\KwIn{The maximum number of inner and outer loop iterations $T_\mr{in}, T_\mr{out}$;}
			\KwOut{\mbox{The dictionary $\AAA \in \RR^{n \times m}$ and assignment weights 
					$\WW \in \RR_+^{n \times m}$;}}
			\caption{ADMM method for group sparse kernelized clustering}
			
			\BlankLine
			Initialize the kernel matrix:  $k_{ij} \ = \ \exp(-\gamma \! \cdot \! dist_{ij}^2)$\;
			Initialize $\AAA$ as a random selection matrix and $t_\mr{out}$ to 0\;    	
			
			\BlankLine    
			\While{$f(\DD,\WW)$ not converged and $t_\mr{out} \leq T_\mr{out}$}{
				
				\BlankLine
				Initialize $\UU$ and $\ZZ$ to all zeros and $t_\mr{in}$ to zero\;
				
				\BlankLine
				\While{$||\WW-\ZZ||_F^2$ not converged and $t_\mr{in} \leq T_\mr{in}$}{	
					
					\BlankLine
					Update $\WW$, $\ZZ$ and $\UU$: \\
					\begin{fleqn}	\begin{flalign*}
						\qquad \WW & \ \gets \ 
						\big(\tr{\AAA}\KK\AAA + \mu \II\big)^{-1}\big(\tr{\AAA}\KK + \mu(\ZZ-\UU)\big); \\
						\qquad \hat{z}_{ij} & \ \gets \ \max\Big\{w_{ij} + u_{ij} - \frac{\lambda_1}{\mu}, \, 0\Big\}, 
						\quad i \leq m, \ j \leq n; \\
						\qquad \zri & \ \gets \ \max\Big\{||\zhri||_2 - \frac{\lambda_2}{\mu}, \, 0\Big\} \cdot 
						\frac{\zhri}{||\zhri||_2}, \quad i \leq m; \\
						\qquad \UU \ & \ \gets \ \UU \ + \ \big(\WW - \ZZ\big);
						\end{flalign*}\end{fleqn}
					$t_\mr{in} \ \gets \ t_\mr{in} + 1$\;
				}
				
				\BlankLine
				Update dictionary until convergence: $ \AAA_{ij} \ \gets \ \AAA_{ij}\dfrac {\left( \KK \tr{\WW}\right) _{ij}} {\left( \KK \AAA \WW \tr{\WW}\right) _{ij}}, \quad i=1,\ldots,n, \quad j=1,\ldots,m.$\;
				$t_\mr{out} \ \gets \ t_\mr{out} + 1$\;
			}	 	 	 	
			
			\BlankLine
			\Return{$\{\AAA, \WW\}$} \;
		\end{algorithm2e}
		\centering
	\end{small}
	
	\subsubsection{Algorithm complexity}
	
	The clustering process of our proposed method is summarized in Algorithm 1. In this algorithm, the user provides a matrix $S_{dist}$ of pairwise streamline distances (see experiments for more details), the maximum number of clusters $m$, as well as the trade-off parameters $\lambda_1, \lambda_2$, and obtains as output the dictionary matrix $\AAA$ and the cluster assignment weights $\WW$. At each iteration, $\WW$, $\ZZ$ and $\UU$ are updated by running at most $T_\mr{in}$ ADMM loops, and are then used to update $\AAA$. This process is repeated until $T_\mr{out}$ iterations have been completed or the cost function $f(\DD,\WW)$ converged. The soft assignment of $\WW$ can be converted to a hard clustering by assigning each streamline $i$ to the bundle $m$ for which $w_{im}$ is maximum. 
	
	The complexity of this algorithm is mainly determined by the initial kernel computation, which takes $O(n^2)$ operations, and updating the assignment weights in each ADMM loop, which has a total complexity in $O(T_\mr{out} \cdot T_\mr{in} \cdot m^2 \cdot n)$. Since $T_\mr{out}$, $T_\mr{in}$ and $m$ are typically much smaller than $n$, the main bottleneck of the method lies in computing the pairwise distances $S_{dist}$ used as input. For datasets having a large number of streamlines (e.g., more than $n=100,000$ streamlines), this matrix could be computed using an approximation strategy such as the the Nystr\"om method \cite{fowlkes2004spectral}, described later in the paper.
	
	\subsection{Kernelized dictionary learning with Laplacian prior}
	
	\subsubsection{Update W: Bartels-Stewart Algorithm summary }
	
	\begin{small}
		\centering
		\begin{algorithm2e}[!htb]
			\KwIn{$\PP$, $\QQ$, and $\RR$;}
			\KwOut{$\WW$}
			\caption{Bartels-Stewart Algorithm summary}
			
			\BlankLine
			Step 1: Transfrom $\PP$ and $\QQ$ into Schur form \\
			\BlankLine
			$\CCC_c \ = \ \RR;$ \\
			$[\QQ_a,\TTT_a] \ = \ schur(\PP); \quad  \CCC_c \ = \ \tr{\QQ_a}\CCC_c$\\
			$[\QQ_b,\TTT_b] \ = \ schur(\QQ); \quad  \CCC_c \ = \ \CCC_c\QQ_b$\\
			
			\BlankLine
			Step 2: Solve following Simplified Sylvester equation using back substitution
			$\TTT_a \WW + \WW \TTT_b \ = \ \CCC_c$
			
			\BlankLine
			Step 3: Recover $\WW$:\\
			$\WW \ = \ \QQ_a \WW \tr{\QQ_b}$

			\Return{$\{ \WW\}$} \;
		\end{algorithm2e}
		\centering
	\end{small}

	\subsubsection{Algorithm summary and complexity}

	The algorithm below provides summary of the methods, while complexity can be computed similar to previous section, with only difference being update of W.

	\begin{small}
		\centering
		\begin{algorithm2e}[!htb]
			\KwIn{Pairwise streamline distance matrix $S_{dist} \in \RR^{n \times n}$;}
			\KwIn{The maximum number of streamline bundles $m$;}
			\KwIn{The RBF kernel parameter $\gamma$;}
			\KwIn{The cost trade-off parameters $\lambda_1, \lambda_L$ ;}
			\KwIn{Lagrangian parameter $\mu_1$;}	
			\KwIn{The maximum number of inner and outer loop iterations $T_\mr{in}, T_\mr{out}$;}
			\KwOut{\mbox{The dictionary $\AAA \in \RR^{n \times m}$ and assignment weights 
					$\WW \in \RR_+^{n \times m}$;}}
			\caption{ADMM method for kernelized dictionary learning with Laplacian prior}
			
			\BlankLine
			Initialize the kernel matrix:  $k_{ij} \ = \ \exp(-\gamma \! \cdot \! dist_{ij}^2)$\;
			Initialize $\AAA$ as a random selection matrix and $t_\mr{out}$ to 0\;    
			Precompute 	$ schur(\lambda_L \LLL)$;
			
			\BlankLine    
			\While{$f(\DD,\WW)$ not converged and $t_\mr{out} \leq T_\mr{out}$}{
				
				\BlankLine
				Initialize $\UU$ and $\ZZ$ to all zeros and $t_\mr{in}$ to zero\;
				
				\BlankLine
				\While{$||\WW-\ZZ||_F^2$ not converged and $t_\mr{in} \leq T_\mr{in}$}{	
					
					\BlankLine
					Update $\WW$, $\ZZ$, and $\UU$: \\
					\begin{fleqn}	\begin{flalign*}
						\qquad \WW & \ \gets \ 
						Sylvester\big(\big(\tr{\AAA}\KK\AAA + \mu_1\II\big),\lambda_L \LLL,\big(\tr{\AAA}\KK + \mu_1(\ZZ-\UU)\big)\big) \\
						\qquad \hat{z}_{ij} & \ \gets \ \max\Big\{w_{ij} + u_{ij} - \frac{\lambda_1}{\mu_1}, \, 0\Big\}, 
						\quad i \leq m, \ j \leq n; \\
						\qquad \UU \ & \ \gets \ \UU \ + \ \big(\WW - \ZZ\big); 
						\end{flalign*}\end{fleqn}
					$t_\mr{in} \ \gets \ t_\mr{in} + 1$\;
				}
				
				\BlankLine
				Update dictionary until convergence: $ \AAA_{ij} \ \gets \ \AAA_{ij}\dfrac {\left( \KK \tr{\WW}\right) _{ij}} {\left( \KK \AAA \WW \tr{\WW}\right) _{ij}}, \quad i=1,\ldots,n, \quad j=1,\ldots,m.$\;
				$t_\mr{out} \ \gets \ t_\mr{out} + 1$\;
			}	 	 	 	
			
			\BlankLine
			\Return{$\{\AAA, \WW\}$} \;
		\end{algorithm2e}
		\centering
	\end{small}

	
	\subsection{Group sparsity and manifold prior visualization}
	
\begin{figure}
		{\centering
			\includegraphics{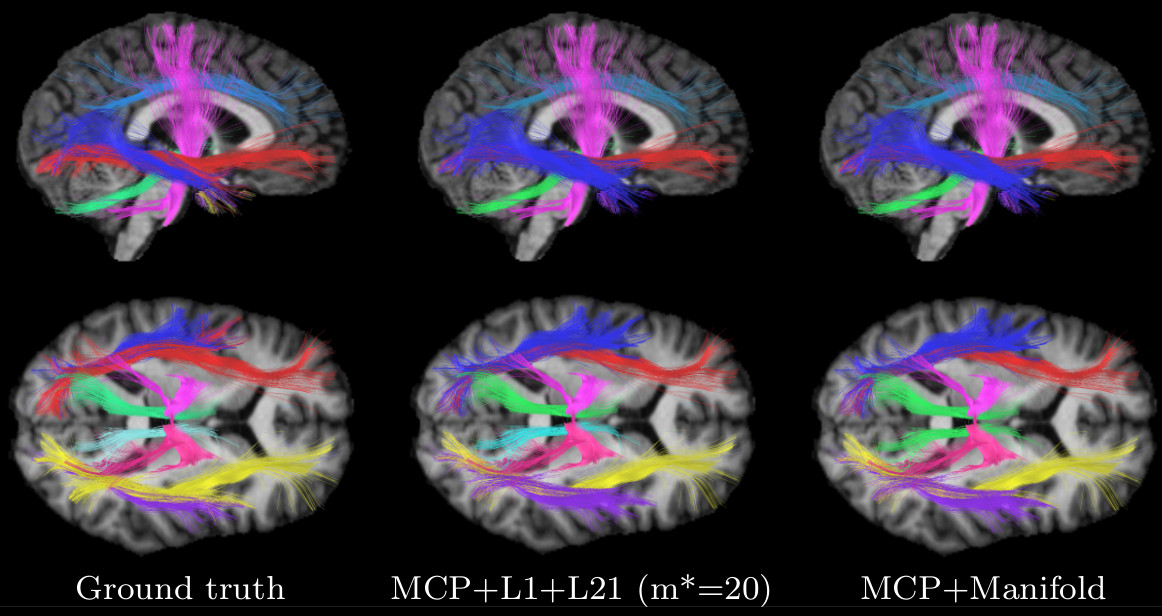}
			
			\vspace{-3mm}
			\caption{Right sagittal (\textbf{top}) and inferior axial (\textbf{bottom}) views of the ground truth (left), and bundles obtained by MCP+L1+L21 (middle, m=20, final m=10), and MCP+L1+Lap (right, m=10).}
			\label{fig:Fiber_bundles_GT_vis_compare_MCP_L21_and_Lap}
		}
	\end{figure}

	The bundles obtained by group sparsity (MCP+L1+L21) for the input number of clusters $m=20$ are presented in Figure \ref{fig:Fiber_bundles_GT_vis_compare_MCP_L21_and_Lap} (middle). We observe that the clustering is similar to the ground truth clustering, except for small differences in left/right inferior longitudinal fasciculus bundles (purple and blue colors in the ground truth). Also, we observe that superior cerebellar peduncle bundles (cyan and green colors in the ground truth) are well separated.
	
	Figure \ref{fig:Fiber_bundles_GT_vis_compare_MCP_L21_and_Lap}(right) shows clustering output using this method for $m=10$ and MCP for a sample run. We observe that the clustering is similar to the ground truth clustering, except for small differences in left/right inferior longitudinal fasciculus bundles (purple and blue colors in the ground truth). 

	
	\subsection{Results on multi-subject MIDAS dataset (KSC+MCP)}
	
	\textbf{Data:} To evaluate the performance of our method on multiple subjects, we also used the data of 12 healthy volunteers (6 males and 6 females, 19 to 35 years of age) from the freely available MIDAS dataset \cite{bullitt2005vessel}. For fiber tracking, we used the tensor deflection method \cite{lazar2003white} with the following parameters: minimum fractional anisotropy of 0.1, minimum streamline length of 100 mm, threshold for streamline deviation angle of 70 degrees. A mean number of 9124 streamlines was generated for the 12 subjects.
	
\begin{figure}[t]
		{\centering    
			\mbox{
				\includegraphics[width=11cm]{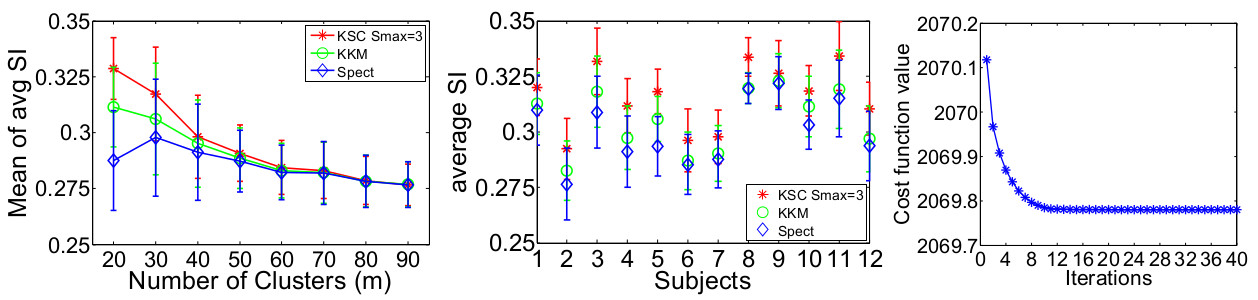}
			}
			\vspace{-3mm}
			\caption{ MIDAS: Mean of average SI using MCP: varying $m$ mean over 12 subjects (\textbf{left}); for $m \ = \ 35$ for each subject (\textbf{middle}); Convergence plot (\textbf{right} for KSC+MCP )}
			\label{fig:MIDAS_Conv_Plots}
		}
	\end{figure}

	\textbf{Results:} Figure \ref{fig:MIDAS_Conv_Plots} (left) shows the mean SI (averaged over all clusters) obtained by KSC ($S_\mr{max}\!=\!3$), KKM and Spect with MCP, on 12 subjects of the MIDAS dataset. We see that our soft clustering method outperforms the hard clustering approaches, especially for a small number of clusters. In Figure \ref{fig:MIDAS_Conv_Plots} (right), the results obtained for $m \ = \ 35$ are detailed for each subject. Error bars in the plot show the mean and variance of SI values obtained over 10 different initializations. As can be seen, our method shows a greater accuracy and less variance across subjects. 
	
	\subsection{Results on Human Connectome Project subjects}
	
\begin{figure}[h]
		{\centering 
			
			\includegraphics{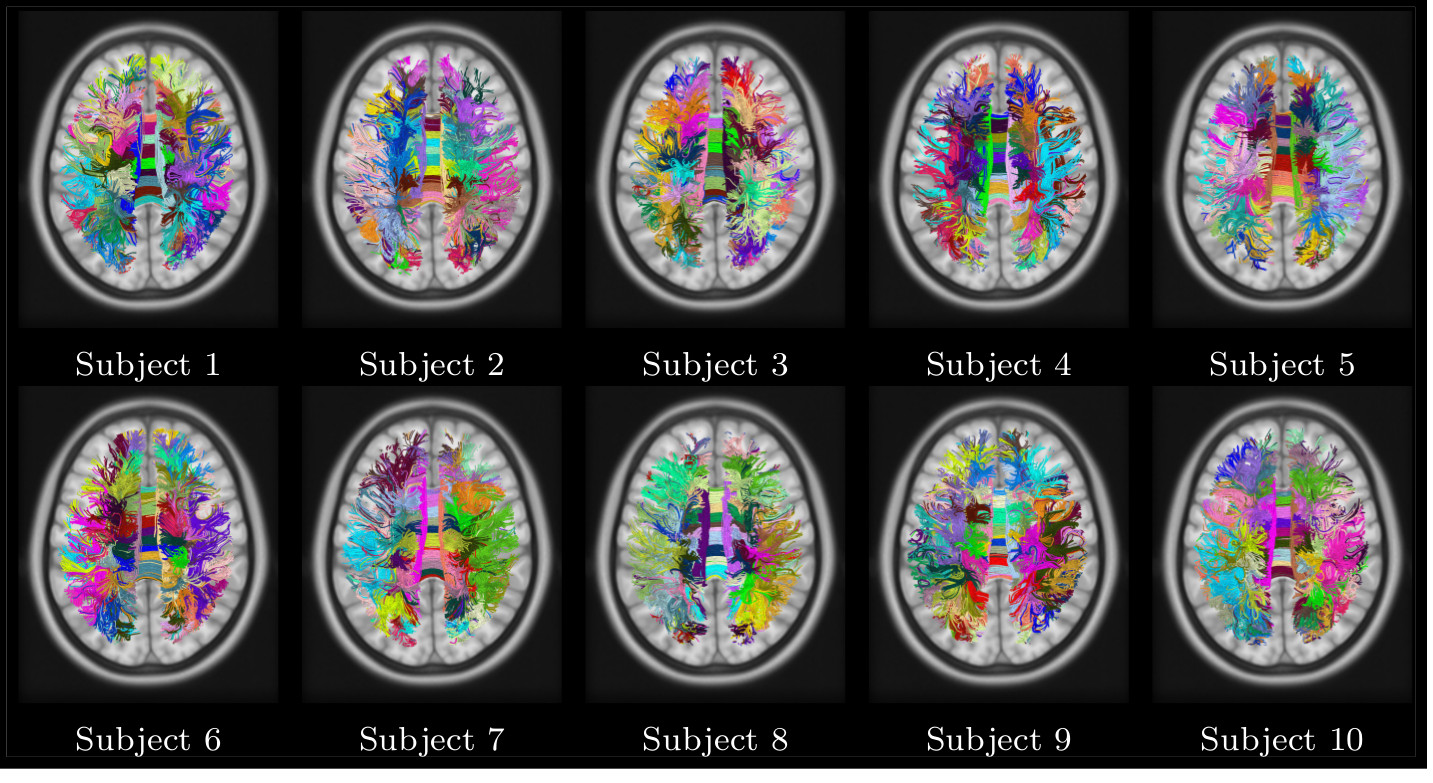}
			
			\vspace{-3mm}
			\caption{Visualization of clustering output for $10$ unrelated HCP subjects using KSC, for m = $50$.}
			\label{fig:HCP10sub_hard_clustering_vis}
		}
	\end{figure}

	Figure \ref{fig:HCP10sub_hard_clustering_vis} shows clustering output for $10$ HCP subjects for $m=50$, with an unique color assigned to each cluster. For this simplified visualization each streamline is assigned to a single cluster by taking the maximum for each column of the matrix $\WW$. Note, we have used a unique color code for each subject, as establishing a cluster correspondence across subjects is itself a challenging problem. We observe that the overall pattern of clustering across subjects looks similar. However, there are subtle variations for clusters across subjects. 
	
\begin{figure}[h]
		{\centering 
			
			\includegraphics{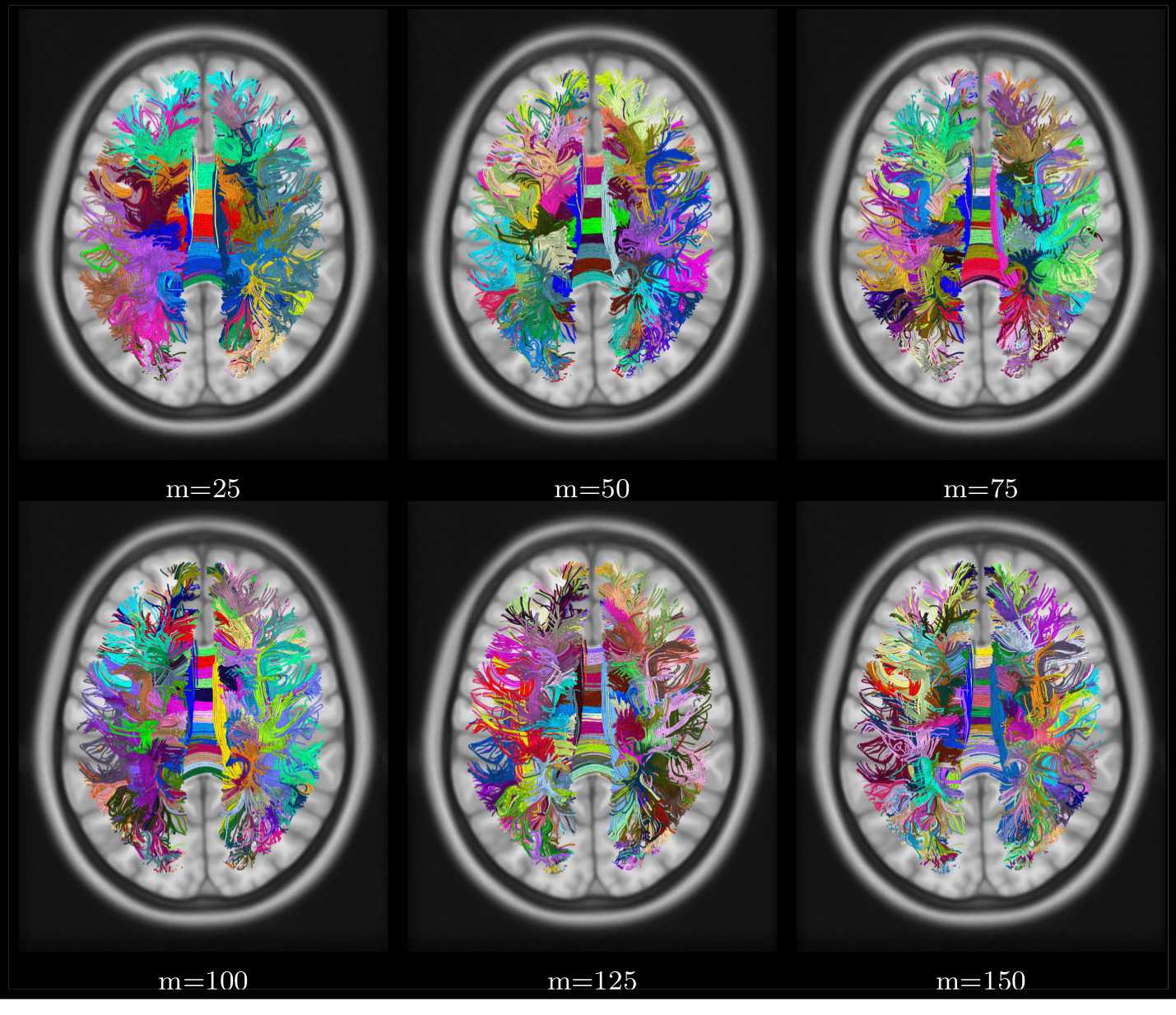}
			
			\vspace{-3mm}
			\caption{Visualization of clustering output for subject 1 using KSC, for varying $m$.}
			\label{fig:Impact_m_HCP_sub1}
		}
	\end{figure}

	Similarly, Figure \ref{fig:Impact_m_HCP_sub1} shows simplified visualization of clustering output for subject 1, for varying $m$. As expected, going from $m=25$ to $m=150$ the clusters split into smaller ones, for example, observe the clusters in corpus callosum.  
	
	\subsection{Multi-subject clustering as dictionary}
		
	\subsubsection{Computing the kernel matrix using the Nystr\"om method}
	
	When there can be multiple subjects, each subject having several thousands of streamlines, computing the similarity between all pairs of training streamlines in $\KK$ is impossible. To alleviate this problem, we approximate the kernel matrix using the Nystr\"om method \cite{fowlkes2004spectral,Odonnell07automatictractography}. In this method, a set of $p$ representative streamlines are sampled from while set of training streamlines, where $p \ll |\XX|$. The pairwise similarities between all selected streamlines are then computed in a reduced kernel matrix $\KK_a \in \RR^{p \times p}$. Likewise, the similarity between the selected and non-selected ones are obtained in a matrix $\KK_b \in \RR^{p \times (|\XX|-p)}$. The whole kernel matrix is then reconstructed using a low-rank approximation $\KK = \GG\tr{\GG}$, where $\tr{\GG} = \KK_a^{-\frac{1}{2}} \big[\tr{\KK}_a \, \tr{\KK}_b\big]$. In practice, the most computationally expensive step of this method is the SVD decomposition of $\KK_a$.
	
	\subsubsection{HCP multi-subject clustering}
	When there are multiple subjects, each subject having several thousands of streamlines, computing the similarity between all pairs of training streamlines in $\KK$ is impossible. To alleviate this problem, we approximate the kernel matrix using the Nystr\"om method \cite{fowlkes2004spectral,Odonnell07automatictractography}. Figure 9 (manuscript) shows simplified visualization of $\AAA$ matrix of 4 sets of $10$ unrelated HCP subjects. (These subjects are utilized as dictionary in next section). We utilized $50,000$ streamlines for each set, sampling $5,000$ streamlines from each subject. 
	
	For simplification, we show full clusterings and select bundles including Anterior Body, and Central Body bundles in Corpus Callosum; Third row: Left Cortico-Spinal-Tract, and Left Arcuate Fasciculus bundles; Last row: Right Cortico-Spinal Tract, and Right Inferior Occipitofrontal Fasciculus bundles. The objective here is to show that the multi-subject clustering output provides plausible clusters, corresponding to well-known anatomical bundles. Also, comparing multi-subject clustering with single subject clustering, we observe overall similarity in terms of clusters, while also reflecting variation. We also observe subtle variations across multi-subject clustering sets, for example, within IOF or CST bundles. 	
	
	
	\subsection{Application: automated segmentation of new subject streamlines}
\begin{figure}[htb!]
		{\centering 
			
			\includegraphics[scale=0.90]{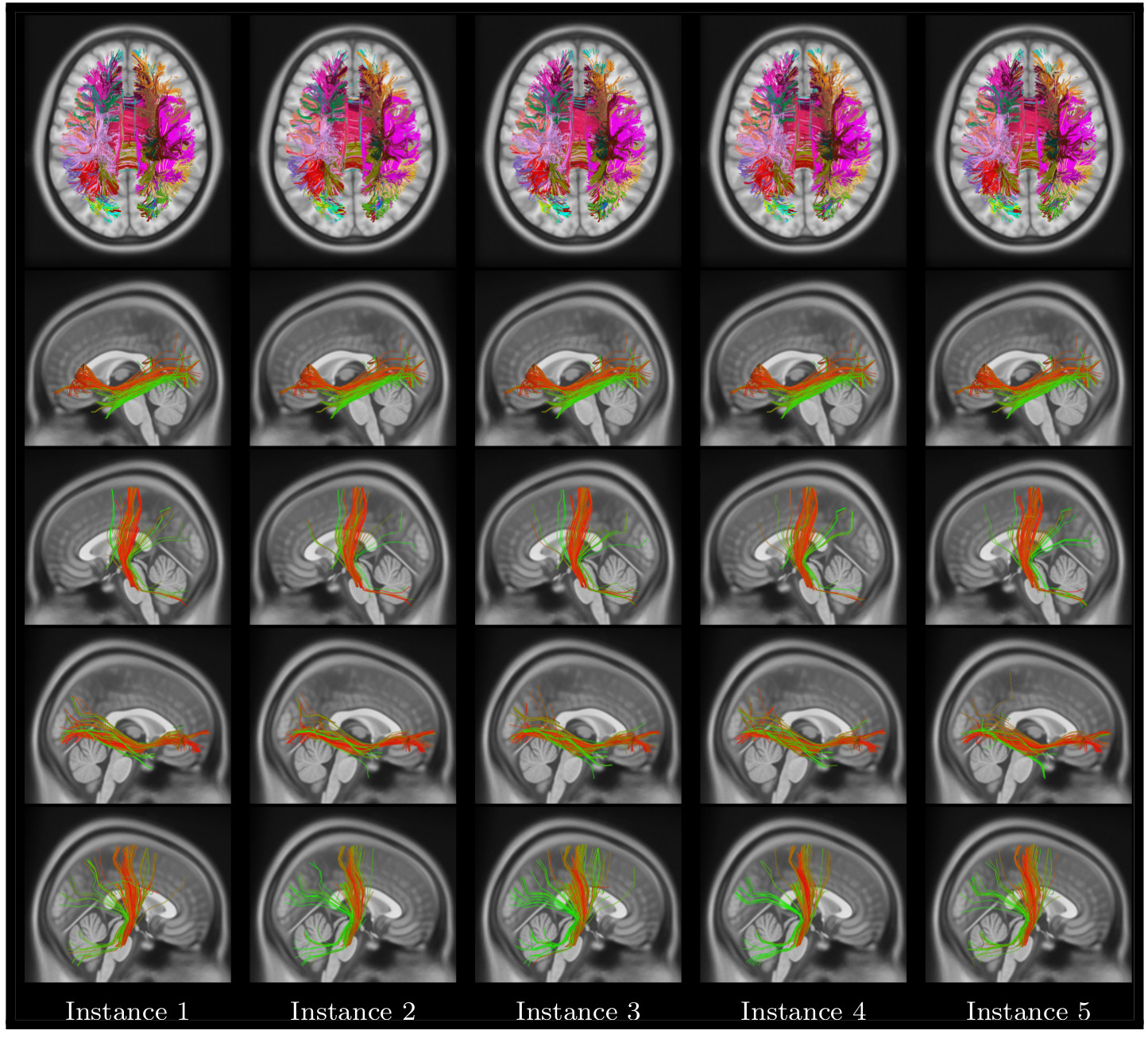}
			
			\vspace{-3mm}
			\caption{Automated segmentation of 5 instances of subject 1 using dictionary D1.}
			\label{fig:Auto_seg_sub2_D4_instance1_to5}
		}
	\end{figure}

	To analyze the impact of sampling streamlines from a subject, Figure \ref{fig:Auto_seg_sub2_D4_instance1_to5} shows segmentation output for 5 instances of subject 1 using dictionary D1. 
	
	\bibliography{References_KD_Sparsity_journal}